\documentclass[USenglish,twocolumn]{article}

\usepackage[utf8]{inputenc}
\usepackage[big]{dgruyter_author}
\usepackage{graphicx}% http://ctan.org/pkg/graphicx
\usepackage{epstopdf}
\usepackage{subfigure}
\usepackage{multicol}
\usepackage{multirow}
\usepackage{flushend}
\usepackage{collcell}
\usepackage{booktabs}
\usepackage[T1]{fontenc}
\usepackage{amsmath}
\usepackage{amsfonts}
\usepackage{comment}
\newcommand{\ignore}[1]{}
\newcommand{\etal}{\textit{et al}. }
  
\begin{document}

  \articletype{Research Article{\hfill}Open Access}

  \author*[1]{Akansel Cosgun}
  \author[2]{Henrik I. Christensen}

  \affil[1]{Monash University, Australia; E-mail: akansel.cosgun@monash.edu}
  \affil[2]{University of California San Diego, USA; E-mail: hichristensen@ucsd.edu}

  \title{\huge Context-aware robot navigation using interactively built semantic maps}

  \runningtitle{Context-aware robot navigation using interactively built semantic maps}

  \begin{abstract}
{We discuss the process of building semantic maps, how to interactively label entities in them, and how to use them to enable context-aware navigation behaviors in human environments. We utilize planar surfaces, such as walls and tables, and static objects, such as door signs, as features for our semantic mapping approach. Users can interactively annotate these features by having the robot follow him/her, entering the label through a mobile app, and performing a pointing gesture toward the landmark of interest. Our gesture-based approach can reliably estimate which object is being pointed at, and detect ambiguous gestures with probabilistic modeling. Our person following method attempts to maximize future utility by search for future actions assuming constant velocity model for the human. We describe a method to extract metric goals from a semantic map landmark and to plan a human aware path that takes into account the personal spaces of people. Finally, we demonstrate context-awareness for person following in two scenarios: interactive labeling and door passing. We believe that future navigation approaches and service robotics applications can be made more effective by further exploiting the structure of human environments.}
\end{abstract}
  \keywords{mobile robot navigation, semantic mapping, human-robot interaction}
%  \classification[PACS]{}
 % \communicated{...}
 % \dedication{...}

  \journalname{Paladyn, J. Behav. Robot.}

  \startpage{1}

\maketitle

\section{Introduction}

\begin{figure}[ht!]
\begin{center}
\centering
\includegraphics[width=0.48\textwidth]{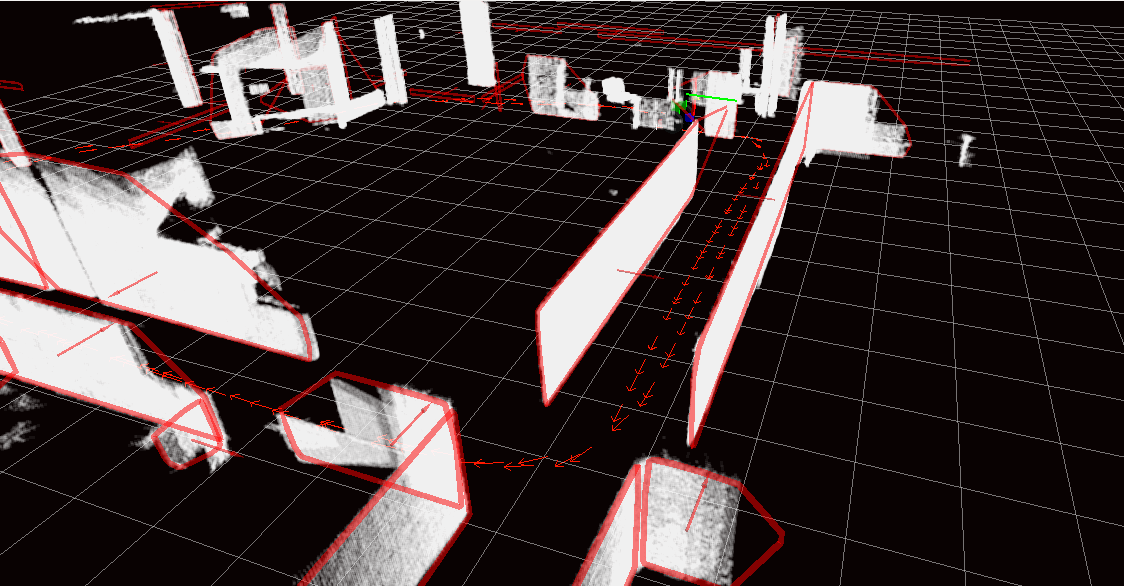}
\caption{An example of the type of map produced by our system. Planar features
are visible by the red convex hulls and red normal vectors. The small red arrows on
the ground plane show the robot's trajectory. The point clouds used to extract these
measurements are shown in white and have been rendered in the map coordinate
frame by making use of the optimized poses from which they were taken.} 
\label{fig:fig1}
\end{center}
\end{figure}

Millions of robots around the world are in operation today, however, most of them operate in factories and are physically separated from humans. In the future, robots could be deployed in human environments, such as hotels, hospitals, offices and homes, and they could be used for elderly care, cleaning, welcoming guests, and object delivery. There are two observations we make for this problem domain. First, human environments are designed for human convenience. For example, rooms offer privacy and compartmentalization of activities, doors are easy to open for humans, people put their things on planar surfaces, and door signs help people distinguish different rooms. Contemporary approaches to robot navigation typically do not take advantage of such human-made structures. Second, robots will be in close proximity to humans and interacting with us on a daily basis. Standard robot path planning algorithms do not distinguish people from obstacles, thus ignoring the social aspect of navigation. In this paper, we focus on how to utilize human-made structures to improve reasoning capabilities of service robots, as well as complying with social conventions to navigate efficiently among people. 

We aim to develop intelligent mobile robots that understand the semantics of human environments and the spatial relationships with and between humans. Our mapping approach leverages our prior knowledge of semi-structured human environments to provide a rich representation for service robotics tasks. Specifically, our maps contain high-level features such as objects, planar surfaces and signs that contain text in addition to metric coordinates. For example, planar landmarks enable the robot to know the locations of tables, counters, rooms, and doors. Door sign landmarks enable room locations and numbers to be automatically added to the map, while object landmarks can be used for fetch and carry tasks. Our approach additionally supports manual annotation of these high-level landmarks, so that they can be referenced by name in interactions with users. 

We use the user-annotated landmarks and people tracking in four ways to enable context-aware robot navigation: First, high-level features, such as planar surfaces, is used for robust localization. Second, users and robots refer to the same landmarks by name, which enable users to provide human-friendly navigation goals instead of goals in metric coordinates. Third, knowledge of nearby landmarks is used to infer the user's intention, which helps the robot to move appropriately in certain tasks such as door passing and person following. Finally, our path planner treats humans differently than obstacles, predicts future trajectories of people and takes into account human safety and comfort.

This paper integrates our previous work on interactive semantic map building and introduces navigation behaviors that use the information contained in the semantic maps. The contributions of this paper are as follows: 
\begin{itemize}
\item A rich map representation that contains high-level features, such as planar surfaces, static objects, and door signs that are grounded in metric coordinates
\item A multi-modal interaction model for annotating semantic landmarks based on natural gestures and an app; and
\item Use of the annotated landmarks and person tracking for demonstrating context-aware navigation behaviors.
\end{itemize}
We make use of external algorithms including GoogleGoggles image recognition engine, GTSAM mapping \cite{dellaert2006square}, and OpenNI NITE skeleton tracker.

The rest of this paper is organized as follows: A literature survey of related work is given in Section \ref{sec:related_works}, followed by our semantic mapping approach in Section \ref{sec:slam_and_semantic_mapping}. Section \ref{sec:interactive_map_labeling} describes how users interactively label semantic elements, as well as sub-components, such as person following and pointing gestures, that make interactive labeling possible. In Section \ref{sec:evaluation}, we evaluate these sub-components. Section \ref{sec:robot_navigation} discusses context-aware navigation behaviors and we conclude in Section \ref{sec:conclusions}.

\section{Related work}
\label{sec:related_works}

Research on semantic mapping and context-aware robot navigation has been ongoing for several years, and a large body of work exists that is related to this paper. 

We first survey mapping techniques for mobile robotics in Section \ref{sec:mapping}, and do a deep dive on semantic mapping in Section \ref{sec:semantic_mapping} and human-augmented mapping in Section \ref{sec:rel_human_augmented_mapping}. We then review literature on person detection and tracking techniques in Section \ref{sec:rel_person_tracking}. Section \ref{sec:rel_context_aware_navigation} is concerned with context-aware navigation where we provide related work on human-aware navigation in Section \ref{sec:human_aware_navigation} and navigation using semantic information in Section \ref{sec:navigation_using_semantics}.

\subsection{Mapping in robotics}
\label{sec:mapping}

For most tasks, mobile robots need to keep a representation of the environment based on sensor readings and possibly prior knowledge. The probabilistic formulation of creating this representation and building a map, called Simultaneous Localization and Mapping (SLAM), was first addressed by Smith and Cheeseman \etal \cite{smith1986representation}, and Leonard and Durrant-Whyte \etal \cite{leonard1991simultaneous}. There are usually two types of SLAM approaches: filter-based and graph-based. Early works used filter-based SLAM approaches that focused on the temporal aspect of the sensor measurements. Graph-based SLAM approaches, instead of solving for only the current robot pose, typically maintain a graph of the entire robot trajectory in addition to the landmark positions \cite{dellaert2006square}. Another area of interest is feature-based SLAM, which uses landmarks to solve the SLAM problem, such as the M-Space model \cite{folkesson2007m}. A central challenge to SLAM is the data association problem, especially when the robot revisits a location \cite{williams2009comparison}.

Three types of map representations are commonly used in robotics: metric, topological, and semantic. Metric maps typically use low-level representations, such as raw sensor measurements (i.e. point clouds \cite{henry2010rgb}), positions of salient features, or occupancy grids \cite{elfes1989using, grisetti2007improved}. In topological maps \cite{remolina2004towards, boal2014topological}, the environment is represented as a graph where nodes represent the distinct places in the environment and edges represent the connections between the places. Semantic maps aims to build richer, more useful maps that include objects, their categories, and common-sense knowledge. 

In our work, we use a hybrid representation: a semantic map for task-level goal assignment and human-robot interaction, and a metric map for motion planning. Below, we review the literature on semantic mapping.

\subsection{Semantic mapping}
\label{sec:semantic_mapping}

Semantic mapping uses high-level modalities such as object recognition, optical character recognition and interaction with humans. Kuipers \etal \cite{kuipers2000spatial} proposed the Spatial Semantic Hierarchy (SSH), which is a qualitative and quantitative model of knowledge of large-scale space consisting of multiple interacting representations. This map also informs the robot of the control strategy that should be used to traverse between locations in the map. Martinez-Mozos \etal \cite{mozos2005supervised} introduce a semantic understanding of the environment creating a conceptual representation referring to functional properties of typical indoor environments. Ekvall \etal \cite{ekvall2007object} integrated an augmented SLAM map with information based on
object recognition, providing a richer representation of the environment in a service
robot scenario. N{\"u}chter \etal \cite{nuchter2008towards} investigated semantic labeling of points in 3D point cloud based maps. Semantic interpretation was given to the resulting maps by labeling points or extracted planes with labels such as floor, wall, ceiling, or door. Pronobis \etal \cite{pronobis2012large} proposed a joint spatial-semantic environment model by fusing multi-modal data including natural language and object classifiers. Recent work in semantic mapping include object-oriented semantic mapping \cite{choudhary2016multi,sunderhauf2017meaningful}. These methods uses objects as landmarks and can create maps that are meaningful to humans. Semantic maps can be useful for describing spatial relations with natural language \cite{fasola2013using,tellex2011understanding}, such as understanding commands like ``get the mug on the table''. For more related work on this topic, the reader is referred in-depth surveys on perception approaches to semantic mapping by Kostavelis \etal \cite{kostavelis2015semantic} and on spatial reasoning by Landsiedel \etal \cite{landsiedel2017review}.

In our approach, we utilize multiple modalities of semantic features, including household objects, door signs and labeled planar surfaces.

\subsection{Human-augmented mapping}
\label{sec:rel_human_augmented_mapping}

Human-augmented mapping was first introduced by Topp \etal \cite{topp2006topological}, 
where a human assists the robot in the map building
process. This is motivated by the scenario of a human guiding a service robot on a
tour of an indoor environment and adding relevant semantic information to the map
throughout the tour for later reference. Users could ask the robot to follow them
throughout the environment and provide labels for locations, which could later be
referenced in commands, such as ``go to label''. This means of providing labels seems
quite intuitive, as users are co-located in the environment with the robot platform.

One of the key concepts in semantic mapping is that of ``grounding'', or establishing 
``common ground'' \cite{clark1991grounding}. Of particular interest for mapping is grounding references,
in order to ensure that the human and robot have common ground when referring
to regions of a map, structures, or objects. Many spatial tasks may require various
terms to be grounded in the map. Dialog in human augmented mapping has been investigated in \cite{kruijff2006clarification}. 
Clarification dialogs were studied in order to resolve ambiguities in the mapping process,
for example, resolving whether or not a door is present in a particular location. This was applied to the Cosy Explorer system, described in \cite{zender2007integrated}, 
which includes a semantic mapping system that build multi-layered maps, including a metric feature based map, a topological map, as well as detected objects. 

Gemignani \etal \cite{gemignani2016interactive} presents evaluation of an interactive semantic mapping system.  In contrast to our work, their approach does not utilize semantic features as landmarks during SLAM. Their approach, however, extracts a topological map from the semantic map in order to facilitate task planning. This work and many others use natural language as the modality to provide labels for the semantic map whereas we use a smartphone app.

\subsection{Person detection and tracking}
\label{sec:rel_person_tracking}

The applicability of person detection and tracking is wide ranging, including congestion analysis in crowded places, security, diagnostics of orthopedic patients, autonomous vehicles and human-computer interfaces. A large body of work exists in the computer vision area; an extensive survey is given in \cite{moeslund2006survey}. Popular methods in image-based person detection include using temporal templates \cite{bobick2001recognition}, histogram-based methods \cite{dalal2005histograms}, deformable part-based methods \cite{mikolajczyk2004human, shu2012part} and multi-modal methods \cite{darrell2000integrated}. Depth cameras are commonly used for body pose estimation \cite{shotton2011real}. More recently, convolutional neural networks \cite{lecun1995convolutional} and deep learning \cite{krizhevsky2012imagenet} became the dominant method for image-based object detection. These techniques has been applied to  person detection \cite{tian2015deep} and tracking \cite{alahi2016social}.

For mobile robotics, laser scanners remain the most commonly used sensor for person detection and tracking, because, as opposed to monocular cameras, they can more easily determine the distance to the detections and their higher field of view makes it possible for a single sensor to cover the surroundings of the robot. Legs in laser scans are typically distinguished using a multitude of geometric features \cite{arras2007using}. 
Schulz \etal \cite{schulz2001tracking} uses particle filters and statistical data association.
Topp \etal \cite{topp2005tracking} demonstrates that leg tracking in cluttered
environments is possible, but prone to false positives. Bellotto \etal \cite{bellotto2009multisensor} combine leg detection 
and face tracking in a multi-modal tracking framework. Zanlungo \etal \cite{zanlungo2011social} utilize the Social Forces Model to describe pedestrian motions, where parameters are trained with real pedestrian data. Leigh \etal \cite{leigh2015person} track multiple people with laser scanners. Dondrup \etal \cite{dondrup2015real} present a framework that utilizes multiple sensor modalities for real-time tracking.

Person tracking provides the robot with the position, and potentially orientation of the humans. However, richer information is typically needed for Human-Robot Interaction (HRI) applications. Pointing gestures are commonly used in HRI, such as for object references \cite{schmidt2008interacting} and providing navigation goals \cite{van2011real}. After deciding if a pointing gesture occurred or not, typically the direction of pointing
is also estimated. A commonly used method is to extend a ray from a body part to another and assume this ray is aimed toward the object of interest. The two of most commonly used methods are elbow-hand \cite{brooks2006working} and head-hand rays \cite{schmidt2008interacting}.

We use a laser-based torso detection approach and track each person individually using a Kalman Filter. Our approach to data association is nearest neighbors. Our pointing gesture approach can take as input both the elbow-hand and head-hand rays, and uses pointing statistic priors to determine the target object.

\subsection{Context-aware navigation}
\label{sec:rel_context_aware_navigation}

Path planning for mobile robotics is traditionally seen as a shortest-path problem and doesn't utilize semantic information. While such approaches generate collision-free paths, the resulting robot behavior may not be preferable to humans. For example, the robot would make people feel unsafe by getting too close to them, or it won't be able to predict the intentions of people if it doesn't recognize gestures. Context-aware navigation has found interest in two fronts: human-aware navigation and navigation using semantic information.

\subsubsection{Human-aware navigation}
\label{sec:human_aware_navigation}

Human-aware navigation algorithms are concerned with planning a motion for a mobile robot given obstacles and people around. 

A common situation in human environments is when the robot encounters bystanders on the way to its goal position. An approach to encode mobility constraints for navigating around humans is through costmaps \cite{sisbot2007human, kirby2009companion}. These approaches typically model personal spaces by assigning costs according to distance and orientation of the robot with respect to humans. Walters and Dautenhahn \etal \cite{walters2005influence} show that people's personal spaces can differ according to their personality, gender and preferences. Predicting the future movements of people is important for planning robot motion. Luber \etal \cite{luber2012socially} trains a model to estimate the future relative motion of people and plan a path. Kidokoro \etal \cite{kidokoro2015simulation} simulates hypothetical situations using real data to anticipate how pedestrians' walking comfort would be affected. Bordallo \etal \cite{bordallo2015counterfactual} and K{\"o}eckemann \etal \cite{koeckemann2015inferring} first explicitly estimate the goal of the people, and then plan for the robot motion accordingly. Understanding the predictability and legibility of robot motion by human observers is a relevant factor in designing robot behaviors \cite{kruse2012legible, dragan2013legibility}. There has been efforts to extend the aforementioned ideas to navigation among crowds \cite{trautman2015robot, henry2010learning}.

Another type of an application is when a person is part of the goal definition, such as when the robot is following, guiding \cite{philippsen2003smooth} or moving alongside a specific person \cite{morales2012people}. Our work involves person following and here we review related works on that topic. Ohya \etal \cite{ohya2002intelligent} present a following method that escorts a target on the side while avoiding obstacles. It was assumed that the target would move with the same acceleration and velocity. Murakami \etal \cite{murakami2014destination} present a method to first estimate the sub-goal of the leading person and then following as if the robot knows the goal. Park \etal \cite{park2013autonomous} model the problem as a control problem and offer an algorithm based on Model Predictive Control. Granata \etal \cite{granata2012framework} present behaviors such as going towards, following and searching a user. Gockley \etal \cite{gockley2007natural} compared two elementary following methods: direction following, where the robot always attempts to drive towards the tracked person, and path following, in which the robot follows the exact path the person took. It was shown that direction following behavior was perceived as more human-like and natural than path following. More detailed surveys on human-aware navigation can be found in \cite{charalampous2017recent, kruse2013human}

We use a costmap-based approach similar to Sisbot \etal \cite{sisbot2007human} and Kirby \etal \cite{kirby2009companion}. Similar to Bordallo \etal \cite{bordallo2015counterfactual}, the path is planned by taking into consideration the future movements of humans. Our person following approach involves a limited-horizon search and allows different robot positioning around the human.

\subsubsection{Navigation using semantic information}
\label{sec:navigation_using_semantics}

The robot can exploit the information contained in semantic maps and possibly prior domain knowledge from the environment to increase the  effectiveness of its navigation capabilities. 

Regier \etal \cite{regier2016foresighted} present a planner that predicts traversal costs of potentials paths by considering the amount of clutter in the environment. Pacchierotti \etal \cite{pacchierotti2005human} adjust the robot's speed when the robot is in a hallway setting. Wilde \etal \cite{wildelearning} learns the cost function weights for path planning from users who choose the path for the robot in a map that contains semantic information. Galindo \etal \cite{galindo2013inferring} generate goals for the robot when there are violations of semantic knowledge.

Natural gestures and spoken language are often used to boost HRI: Lu \etal \cite{lu2013towards} show that using gaze cues makes robot-human hallway passing more efficient. Loper \etal \cite{loper2009mobile} presents a system that is capable of responding to verbal and non-verbal gestures and following a person. Anderson \etal \cite{anderson2017vision} present a method that interprets visually-guided navigation instructions using deep learning. Tellex \etal \cite{tellex2011understanding} address the same problem but use a graphical model.

Zender \etal \cite{zender2007human} considers context-awareness for person following, specifically for handling of door and corridor passages. To handle door passages, the robot increases its following distance and that leads the robot to wait for a while. Our approach to navigation using semantics is similar to this work, as we also use objects such as doors for case-based behavior generation.

\section{Semantic mapping}
\label{sec:slam_and_semantic_mapping}

As service robots become increasingly capable and are able to perform a wider
variety of tasks, we believe that new mapping systems could be developed to better
support these tasks. Towards this end, we developed a SLAM approach that uses planar surfaces and objects as landmarks, and maps their locations and extent. We chose planar surfaces because they are prevalent in indoor environments in the forms of walls, tables, and other surfaces. We also utilize door signs, and use this information to enhance robot navigation behavior.

Non-technical users prefer human terms
for objects and locations when assigning tasks to robots instead of whatever indices or coordinates the robot uses to represent them in its memory. Semantic mapping offers an advantage for robots to understand task assignments given to them by human users. We allow humans to label planar landmarks that are automatically acquired during the SLAM process, as described in Section \ref{sec:interactive_map_labeling}. Users provide navigation goals in terms of these labeled landmarks. Our approach of finding goal points for a given planar landmark will be discussed in Section \ref{sec:finding_the_goal_point}.

Planar landmarks provide semantic information about the space, as vertical planes correspond to walls, showing how space is partitioned, while horizontal planes correspond to tables and shelves, where objects of interest may occur. We describe in Section \ref{sec:door_signs} how higher level objects, specifically door signs, can be used as landmarks in SLAM.  We further explore in Section \ref{sec:following_door_passing} how detection of door signs and therefore the existence of doors, can be used for robot navigation.

\subsection{Plane landmarks}

We believe that feature-based maps are suitable for containing task-relevant information for service robots. For example, a home service robot might need to know the locations of kitchen tables, countertops, cupboards and shelves. Structures such as walls could be used to better understand how space is
structured and partitioned. We describe a SLAM approach capable of creating maps of
the locations and extents of planar surfaces in the environment using both 3D and 2D landmarks.

Our SLAM implementation makes use of the GTSAM library \cite{dellaert2006square}. This library
represents the graph SLAM problem with a factor graph which relates landmarks to
robot poses through factors. GTSAM builds a factor graph of nonlinear measurements. Our approach involves using multiple types of landmark measurements as factors of nonlinear measurements. Planar surfaces are detected in point cloud data generated by a Asus Xtion RGB-D camera. An example of a map produced by our system is shown in Figure \ref{fig:fig1}.

A plane in $\mathbb{R}^3$ has the equation
\begin{equation} ax + by + cz + d = 0, \end{equation} where $a$, $b$, $c$, $d$ are parameters that define the plane and $x$, $y$, $z$ are cartesian coordinates of a point that lies on the plane. We use this representation for planes, while additionally representing the plane's extent by calculating the convex hull of the observed points. While only
the plane normal and perpendicular distance are used to correct the robot trajectory in SLAM, it is essential to keep track of the extent of planar patches, as many
coplanar surfaces can exist in indoor environments, and we would like to represent
these as distinct entities. We therefore represent planes as
\begin{equation}
p = [n, hull],
\end{equation}

where
\begin{equation}
n = [a, b, c, d]
\end{equation}
and \textit{hull} is a point cloud consisting of the vertices of the plane's convex hull. As
planes are re-observed, their hulls are extended with the hull observed in the new
measurements. That is, the measured hull is projected onto the newly optimized
landmark's plane using its normal, and a new convex hull is calculated for the sum
of the vertices in the landmark hull and the measurement's projected hull. In this
way, the convex hull of a landmark can grow as additional portions of the plane are
observed.

We use a Joint Compatibility Branch and Bound (JCBB) technique for data association \cite{neira2001data}. JCBB works by evaluating the joint
probability over the set of interpretation trees of the measurements seen by the robot
at one pose. The output of the algorithm is the most likely interpretation tree for
the set of measurements. We are able to evaluate the probability of an interpretation
tree quickly by marginalizing out the irrelevant portions of the graph of poses and features. The branch and bound recursion structure from the EKF formulation is
used in our implementation.

Given a robot pose $X_r$, a transform from the map frame to the robot frame in the form of $(R, \vec{t})$, a previously
observed feature in the map frame $(\vec{n}, d)$ and a measured plane $(\vec{n}_m , d_m )$, the measurement function $h$ is

\begin{equation}
h=\bigg(
\begin{array}{c}
R^T*\vec{n}\\
\ \langle \vec{n},t\rangle + d \\
\end{array}
\bigg) -
\bigg(
\begin{array}{c}
\vec{n}_m\\
\ d_m\\
\end{array}
\bigg).
\end{equation}

The Jacobian with respect to the robot pose is

\begin{align}
\frac{\partial h}{\partial X_r}
 =
\begin{bmatrix} 
 0 & -n_a & n_b & 0\\
 n_a & 0 & -n_c & 0\\
 -n_b & n_c & 0 & 0\\
 0 & 0 & 0 & \vec{n}^T
\end{bmatrix}
.
\end{align}

The Jacobian with respect to the landmark is

\begin{align}
\frac{\partial h}{\partial n_{map}}
 =
\begin{bmatrix} 
 [R_r] & \vec{0}\\
 \vec{X_r^T} & 1
\end{bmatrix}
.
\end{align}

Using this measurement function and its associated Jacobians, we can utilize planar normals and perpendicular distances as landmarks in our SLAM approach. During
optimization, the landmark poses and robot trajectory are optimized.

\subsection{Object landmarks: door signs}
\label{sec:door_signs}

The previous section introduced how we use planar landmarks for SLAM. In this section, we present a method for using a learned object classifier in a SLAM context to provide measurements suitable for mapping.

First, walls are extracted from straight lines in the laser scan. We use a
RANSAC technique to extract lines from the laser data. Only lines which are longer than a certain threshold are passed to the mapper as measurements.

\begin{figure}[ht!]
\begin{center}
\centering
\includegraphics[width=0.3\textwidth]{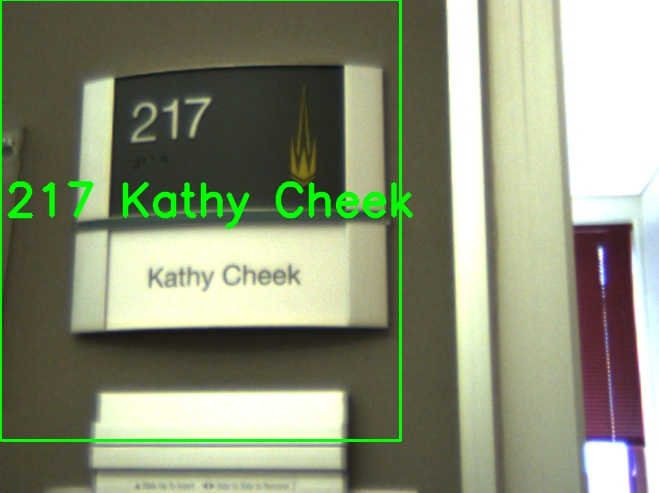}
\caption{This sign is recognized and a measurement is made in the mapper. GoogleGoggles has read both the room number and the text, so this sign can be used for
data association.} 
\label{fig:fig2}
\end{center}
\end{figure}

The door-sign-detector module makes use of a Support Vector Machine (SVM) classifier, trained on Histogram of Oriented
Gradient (HOG) features. If an image region
is classified as a sign by the SVM then a query is made from this image region to
the GoogleGoggles server. If GoogleGoggles is able to read any text on the sign then
it will be returned to us in a response packet. Detected signs with decoded text are
then published as measurements that can be used by the mapper. The measurements
consist of the pixel location in the image of the detected region's centroid, the image patch corresponding to the detected region, and the text string returned from
GoogleGoggles. An example detection of a door sign is shown in Figure \ref{fig:fig2}.

Detected lines in the laser scan and by the door sign detector are added as non-linear measurements to the factor graph. At the time of this study, we did not have a RGB-D sensor on the robot. Therefore, measurements were made on the 3D coordinates of the back-projected image location directly. Range is recovered by finding the laser beam from the head laser which
projects most closely to the image coordinates of the sign. This technique approximates the true range. This factor also incorporates an additional variable which corresponds
to the transformation between the robot base and the camera. 

To implement this factor in GTSAM, we must specify an error function and the
error function's derivatives in terms of all of the variables which contribute to it. The
error function is the difference in the 3D position of the predicted location of the sign
and the measured value given by the recognition module. 

\section{Interactive map labeling}
\label{sec:interactive_map_labeling}

There are several methods to support the annotation of entities in a robot map. For example, while the robot is building its representation of the environment, it can recognize objects or landmarks, such as doors, tables, rooms, and automatically add these features to its map. Even though such a system would be useful, it may wrongly label some objects. In that case, the correct label can be provided by a human with an interactive system. Custom labels would also allow custom annotations such as ``Joe's Room''.

\begin{figure*}[ht!]
\begin{center}
\centering
\includegraphics[width=\textwidth]{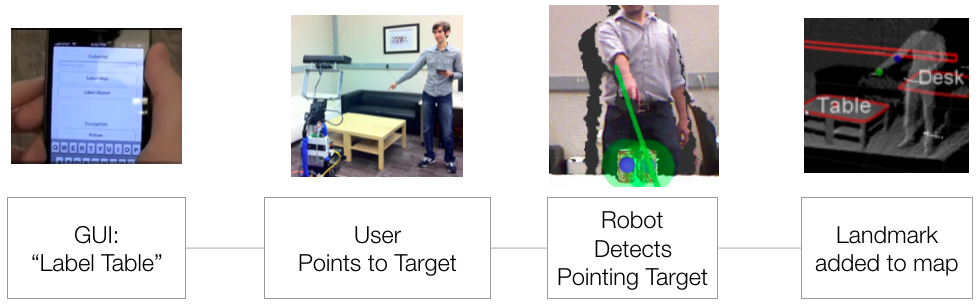}
\caption{Steps for interactively labeling a landmark in the semantic map. First, the user activates person following using the app. The user 
stops nearby the target landmark, and enters the requested label using the app. Then the user performs a pointing gesture towards the target and waits for acknowledgement. The robot assesses the likelihood of nearby objects being the intended target, and asks for confirmation if ambiguity is detected. Finally, a string label is attached to the corresponding landmark in the semantic map.} 
\label{fig:fig3}
\end{center}
\end{figure*}

We presented our method for building semantic maps in Section \ref{sec:slam_and_semantic_mapping}. We will assume that we have a metric and semantic map for the rest of this paper for simplicity. To enable a common ground between the humans and the robot, we developed an interactive procedure to annotate landmarks in the semantic map. In this procedure, the person guides the robot to the landmark of interest first and refers to the landmark of interest by pointing at it. The steps for labeling a landmark is shown in Figure \ref{fig:fig3}. The robot follows the user around between labeling of landmarks. This allows the user to guide the robot to virtually any location in the environment. Our interactive map labeling approach was previously described in \cite{trevor2012interactive}.

Various modalities could be used for the interaction model, such as speech or GUI-only. The reason why we combine the GUI with pointing gestures is that we think natural gestures would play an important role for HRI in the future.

For the rest of this section, we describe the sub-components necessary to realize interactive labeling, namely person following (Section \ref{sec:person_following}), pointing gestures (Section \ref{sec:pointing_gestures}), and object labeling (Section \ref{sec:object_labeling}).

\subsection{Person following}
\label{sec:person_following}

The robot has to continuously estimate the position of the user in real-time for robust person following. We focus on tracking people who are either walking or standing, as these are the two most common human poses around a mobile robot. Below are the brief descriptions of the person detection methods used on the robot as shown in Figure \ref{fig:fig6b}:

1) Leg Detector: A front-facing laser scanner at ankle height (Hokuyo UTM 30-LX) is used. We trained a leg classifier using three geometric features: width, circularity and inscribed angle variance \cite{xavier2005fast}. We find a distance score for each candidate segment using the weighted sum of the distance to each feature and then threshold the score for detection.

2) Torso Detector: A back-facing Hokuyo laser scanner placed at torso-level
is used for this detector. We model
the human torso as an ellipse and fit each segment in the laser as an ellipse. The ellipse fitting approach always returns a result, even for bad data. In addition to the geometric features we use for legs, we use two additional features for torso detection: the horizontal and vertical axes of the fitted ellipse. Similar to leg detection, we use a threshold test for the detection result.

The output of the detectors are input to a state estimation module. Using a state predictor for human movement have two advantages. First, the predicted trajectories are smoother than
raw detections. Smooth tracking helps the robot maintain consistent trajectories
for person following. Second, it provides a posterior estimate that can be used for data association when there is a lack of
matching detections. This allows the tracker to handle temporary occlusions. We use
a linear Kalman Filter (KF) with a constant velocity model to estimate the position and velocity of a person. We used a KF for tracking because it has acceptable tracking performance and is computationally cheap, which is important in real-time applications.

For person following, the robot uses the most probable location of the KF, which is the mean of the Gaussian distribution. We use Dynamic Window Approach (DWA) \cite{fox1997dynamic} at the core of our planner to sample velocity and acceleration-bounded trajectories with the modification of using time as an additional dimension. DWA forward-simulates allowable velocities and chooses an action that optimizes a function that will create a goal-directed behavior while avoiding obstacles. Our approach projects the future locations of the target, creates a tree of trajectories, and scores each tree node according to a goal function which is a function of the relative pose of the robot with respect to the human. Our planner takes the laser scan measurement, predicted positions of the person, and the number of time steps to plan as input and outputs a sequence of actions. A robot configuration at time $t$ is expressed as \(q^t=(x^t,y^t,\theta^t,v^t,\omega^t)^\mathrm{T}\), where $x^t$ and $y^t$ denote positions, $\theta^t$ is the orientation, $v^t$ and $w^t$ are the linear and angular velocities at time $t$. The person configuration \(p^t\) is defined the same way. An action of the robot is defined as a velocity command for some duration: $a(t,\Delta t)=(v^t_a,\omega^t_a,\Delta t)$. We assume a unicycle kinematics model for trajectory sampling. Using this model, we generate a tree up to a fixed depth, starting from the current configuration of the robot. A tree node consists of a robot configuration as well as the information about the previous action and parent node. Every depth of the tree corresponds to a discretized time slice. Therefore, every action taken in the planning phase advances the time by a fixed amount. This enables the planner to consider future steps of the person and simulate what is likely to happen in the future. The planner uses depth-limited Breadth First Search (BFS) to search all the trajectories in the generated tree and determines the trajectory that will give the robot the maximum utility over a fixed time in the future. Given the robot and person configuration at some particular time, goal function \(0 \leq g(q^t,p^t) \leq 1\) determines how desirable the situation is for the task. Goal function can be defined in any way and provides flexibility to the navigation behavior designer. We assume that it is desirable for the robot to follow from behind as it would give the robot a better chance to predict the human's motions and implicitly mimic human's path, which is known to be obstacle free. We report our results on this person following method in Section \ref{sec:eval_person_following}. In previous work, we applied this following method on an autonomous telepresence robot \cite{cosgun2013autonomous}. 

Another important capability to enable interactive labeling is to be able to refer to the same landmarks in the environment. Our design involves the person extending his/her arm and point at the landmark of interest. The method for detecting the pointing target is described next.

\subsection{Pointing gestures}
\label{sec:pointing_gestures}

In this section, we present an uncertainty model for estimating pointing gesture targets based on previous work \cite{cosgun2015did}. Estimating this uncertainty allows us to interpret whether a pointing gesture is ambiguous or not and when objects are too close to one another. We model the uncertainty of pointing gestures using a spherical coordinate system. We use this model to determine the correct pointing target and detect when there is ambiguity. As reviewed in Section \ref{sec:rel_person_tracking}, a common method in inferring pointing gesture directions is to extend a virtual ray from a body part to another. We evaluate two of the most commonly used rays, elbow-hand and head-hand, using a 3rd party skeleton tracking algorithm, OpenNI NITE in Section \ref{sec:eval_pointing_gestures}. We use a simple gesture detection algorithm as our focus is to estimate the gesture target given that a pointing gesture was performed. Using the skeleton data, pointing gestures are recognized if a human's forearm makes more than a fixed angle with the vertical axis and elbow and hand joints stay almost stationary for a fixed duration. Gesture detection is activated only after the user requests a labeling action. This design is intended to reduce false positive detections.

\begin{figure}[h!]
\centering
\includegraphics[width=85mm]{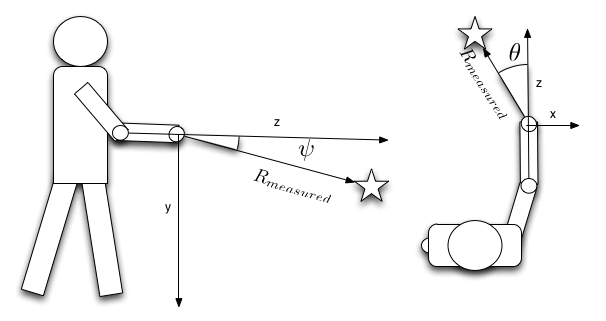}
\caption{Vertical $(\psi)$ and horizontal $(\theta)$ angles in spherical coordinates are illustrated. A potential intended target is shown as a star. The z-axis of the hand coordinate frame is defined by either the Elbow-Hand (this example) or Head-Hand ray.}
\label{fig:fig4}
\end{figure}

We represent a pointing ray in two angles: a ``horizontal'' sense we denote as $\theta$ and a ``vertical'' sense we denote as $\psi$. We first attach a coordinate frame to the hand point, with its z-axis oriented in either Elbow-hand $\vec{v}_{eh}$ or Head-Hand $\vec{v}_{hh}$ directions. The hand was chosen as the origin for this coordinate system because both of head-hand and elbow-hand pointing methods include the user's hand. The transformation between the sensor frame and the hand frame $^{sensor}T_{hand}$ is calculated by using an angle-axis rotation method. An illustration of the hand coordinate frame for Elbow-Hand method and corresponding angles are shown graphically in Figure~\ref{fig:fig4}.

Given this coordinate frame and a potential target point P, we first transform it to the hand frame
\begin{equation}
^{hand}p_{target} = T_{hand} \: p_{target}.
\end{equation}

We calculate the horizontal and vertical angles for the target point $^{hand}p_{target} = (x_{targ}, y_{targ}, z_{targ})$ as

\begin{equation}
[\theta_{target}\;\psi_{target}]=[atan2(x_{targ}, z_{targ})\;atan2(y_{targ}, z_{targ})]
\end{equation}
where $atan2(y,x)$ is a function that returns the value of the angle $\arctan(\frac{y}{x})$ with the correct sign.

We estimate the likelihood of objects being the target using statistical data from previous pointing gesture observations. We observed that head-hand and elbow-hand methods returned different angle errors depending on the target location. Our approach relies on finding error statistics of these approaches, and compensating the error when the target object is searched for. First, given a set of prior pointing observations, we calculate the mean and variance of the vertical and horizontal angle errors for each pointing method. This analysis will be presented in Section \ref{sec:eval_pointing_gestures}. Given an input gesture, we apply correction to the pointing direction and find the Mahalanobis distance to each object in the scene.

When a pointing gesture is recognized and the angle pair $[\theta_{target}\;\psi_{target}]$ is found then a correction is applied by subtracting the mean terms from measured angles
\begin{equation}
[\theta_{cor}\;\;\psi_{cor}]=[\theta_{target}-\mu_{\theta}\;\;\;\psi_{target}-\mu_{\psi}].
\end{equation}
 
We also compute a covariance matrix for angle errors in this spherical coordinate system: 
\begin{equation}
S_{type} = \begin{bmatrix}
\sigma_{\theta}&0\\ 0&\sigma_{\psi}
\end{bmatrix}
.
\end{equation}

We get the values for $\mu_{\theta}, \mu_{\psi}, \sigma_{\theta} ,\sigma_{\psi}$ from Table~\ref{table:pointing_error_stats} for the corresponding gesture type and closest target location. We then compute the Mahalanobis distance to the target
\begin{equation}
D_{mah}(target,method)=\sqrt{ [\theta_{cor}\;\psi_{cor}]^T S_{method}^{-1} [\theta_{cor}\;\psi_{cor}]}
\end{equation} 
 
We use $D_{mah}$ to estimate which target or object is intended. We consider two use cases: the objects are represented as a point or a point cloud. For point targets, we first filter out targets that have a Mahalanobis distance larger than a threshold $D_{mah} > D_{thresh}$. If none of the targets has a $D_{mah}$ lower than the threshold then we decide the user did not point to any targets. If there are multiple targets that has $D_{mah} \leq D_{thresh}$ then we determine ambiguity by employing a ratio test. The ratio of the least $D_{mah}$ and the second-least $D_{mah}$ among all targets is compared with a threshold to determine if there is ambiguity. If the ratio is higher than a threshold then the robot can resort to additional action, such as initiating a dialogue to ask or confirm the intended object.

\subsection{Labeling object models}
\label{sec:object_labeling}

Our method supports labeling two types of landmarks: planar surfaces and objects. The UI shows two labeling buttons ``Label Object'' and ``Label Planar Surface'', so that the robot knows what the user is intending to label. Section \ref{sec:pointing_gestures} demonstrated how point targets or point cloud targets can be referenced via pointing gestures. For labeling planar surfaces, once the pointing gesture is performed, we check whether any planar feature in the semantic map intersects with the corrected gesture direction. For labeling object models, first the large planar surface corresponding to the table is detected. This
is removed from the point cloud, and point clusters above
this are detected. The cluster with a centroid nearest to the
reference point is selected as the object to be modeled. The
cluster’s points are projected into the camera image and
are used to generate a region of interest. SURF features are
detected for the region of interest, and are stored as an object
model along with the provided label.

We assume that the objects are unique and will remain static throughout, which is obviously a strong assumption for real operation. The object consistency problem is tackled by object-based mapping research \cite{choudhary2016multi,sunderhauf2017meaningful}, but is not the focus of this paper.

\begin{figure}[ht!]
\centering
        \subfigure[]{%
            \includegraphics[width=0.98\columnwidth]{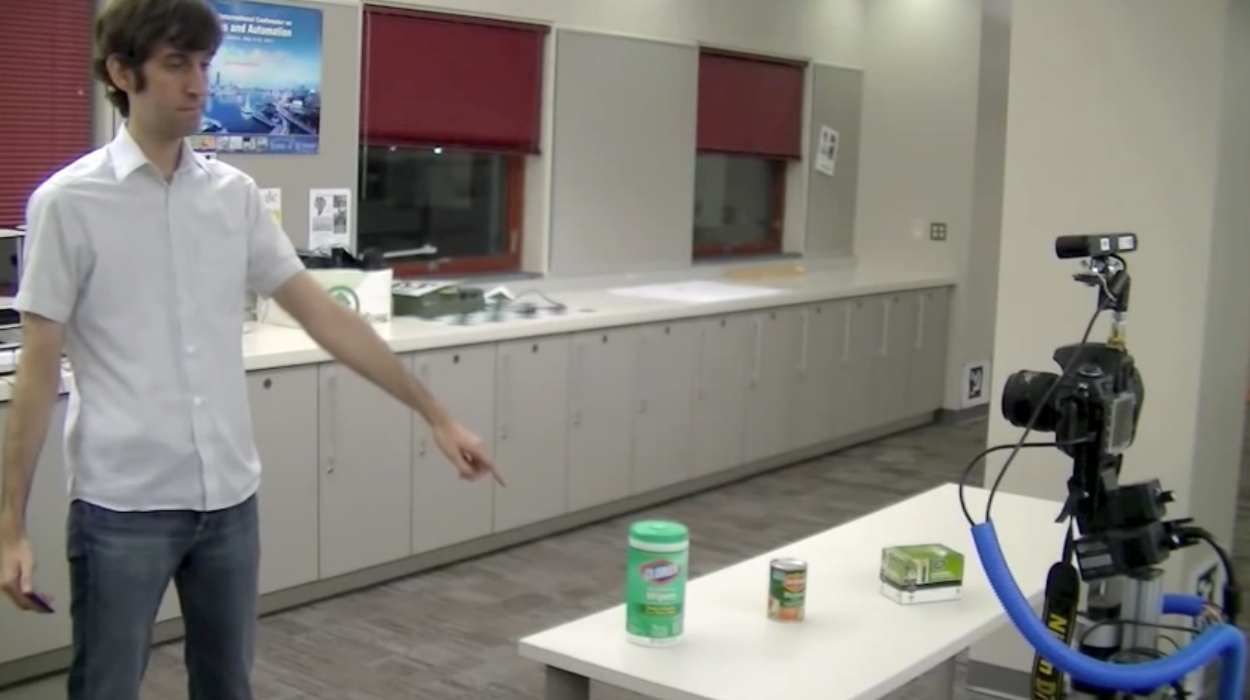}
        }%
        \\
        \subfigure[]{%
           \includegraphics[width=0.6\columnwidth]{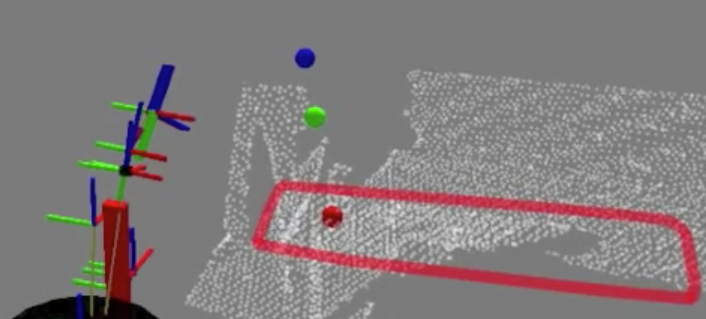}
        }    
         \subfigure[]{%
           \includegraphics[width=0.3\columnwidth]{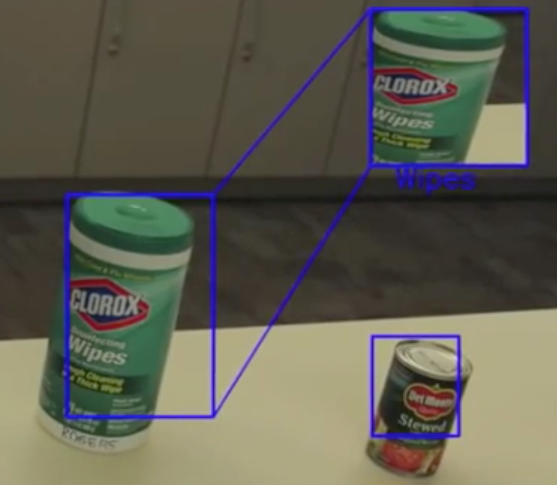}
        }    
    \caption{%
	a) A user pointing at an object; b) A detected pointing gesture (blue and green spheres) and the referenced object (red sphere); c) Features are extracted from an image patch corresponding to the cluster and annotated with the provided label.
     }%
   \label{fig:fig5}
\end{figure}

Once the intended landmark is determined it is annotated
with the label entered by the user and can then be recognized later as described in our previous work \cite{trevor2013interactive}. Figure \ref{fig:fig5} shows the steps the robot executes for this task. Service robotic tasks that use such a map can then reference the object
by label, rather than generating a more complex referring expression (e.g. ``the large object'' or ``the object on the left'').

\section{Evaluation of sub-components}
\label{sec:evaluation}

In this section, we evaluate two of the core sub-components that enable building interactive semantic maps. In Section \ref{sec:eval_person_following} we analyze the person following behavior and in Section \ref{sec:eval_pointing_gestures} we evaluate our pointing gesture target estimation method.

\begin{figure*}[ht]
\centering
        \subfigure[]{%
            \includegraphics[width=0.6\columnwidth]{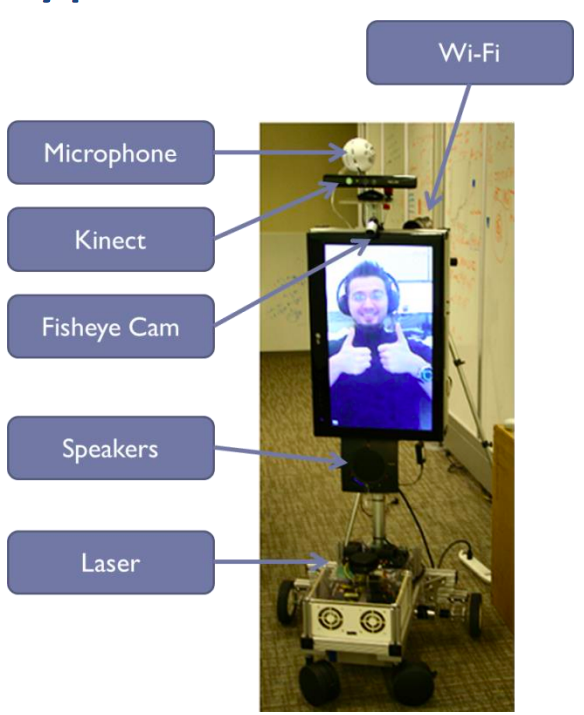}
            \label{fig:fig6a}
        }%
        \subfigure[]{%
           \includegraphics[width=0.51\columnwidth]{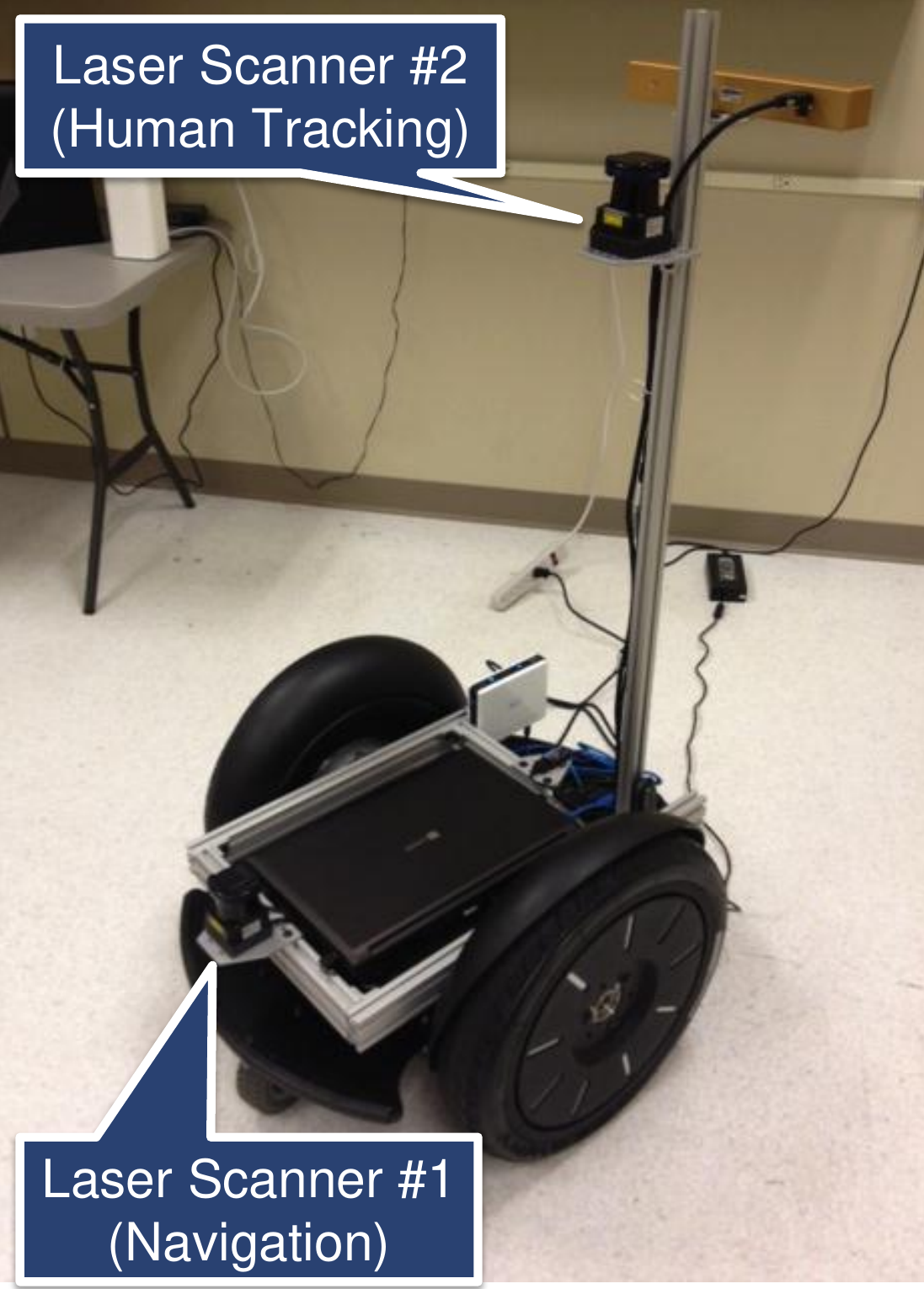}
           \label{fig:fig6b}
        }    
         \subfigure[]{%
           \includegraphics[width=0.6\columnwidth]{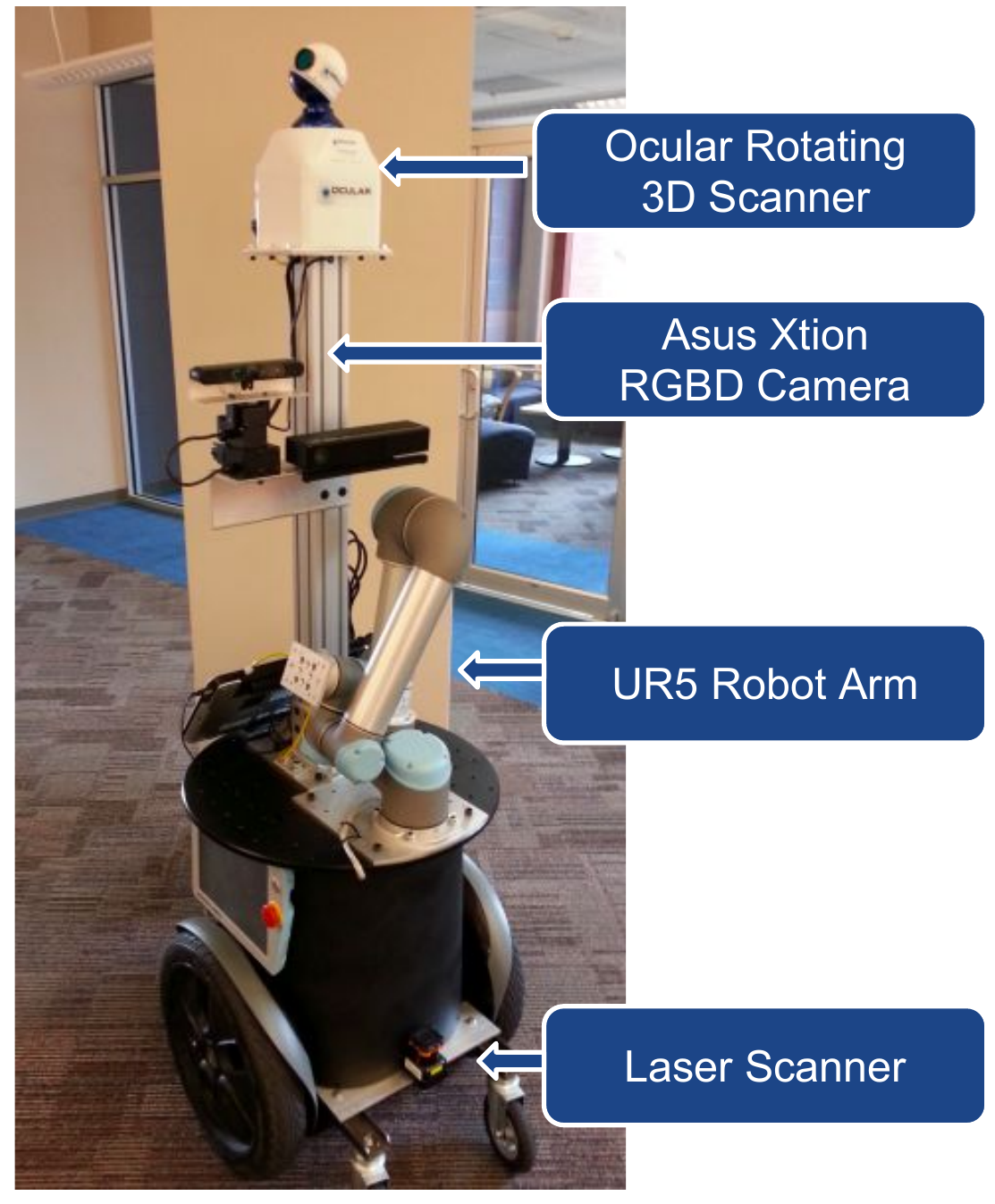}
           \label{fig:fig6c}
        }    
    \caption{%
	Robot platforms used in this paper. The robot shown in a) is a telepresence robot fitted with a Microsoft Kinect sensor. It was used for the person following experiments presented in Figure \ref{sec:eval_person_following}. The robot shown in b) has a Segway base with caster wheels, two laser scanners (one at torso height), and is used in human-aware path planning in Section \ref{sec:human_aware_path_planning}. The robot shown in c) has a Segway base with caster wheels, has a UR5 robot arm, Ocular 3D rotating sensor and an Asus Xtion RGB-D sensor. This robot is used for the rest of the experiments including pointing gestures in Section  \ref{sec:eval_pointing_gestures} and context-aware person following in Section \ref{sec:context_aware_person_following}. All robot platforms have laser scanners for navigation.
     }%
   \label{fig:fig6}
\end{figure*}

The algorithms presented in this paper were implemented on three robot platforms, as shown in Figure \ref{fig:fig6}.

\subsection{Evaluation of person following}
\label{sec:eval_person_following}

In this section, we report on an analysis of the person following sub-component. In our experiments, the robot followed seven people for three laps and we logged the total distance robot followed the person and the average distance to the person. Users did not have prior experience with the robot and were asked to walk in a corridor while the robot is following them. Subjects were encouraged to adjust to the robot's speed which was slightly lower than regular walking speed. A lap consisted of leaving the starting point, going to an intermediate point at the end of the corridor, and coming back to the starting point from the same path. The corridor was L-shaped and did not have any clutter or obstacles other than a couple of tall columns. Occasionally, the robot lost track of people due to fast motions of people, and the experiment was continued after re-initializing the tracking. The goal function for the robot was chosen such that there were two global minimums: a region about $1$m behind the human, and another region that $1$m behind and $0.8$m to the right of the human. These goal regions are designed to encourage the robot to either stay behind the human or shift a bit to one side accordingly to the obstacles around. Table \ref{tab:perf} shows the data pertaining to this study. In total, the robot followed people for about $1.2$km. The robot did not come into contact with any obstacles or people during the experiments and the average distance between the person and the robot across all runs was $1.16$m. 

\begin{table}[!h]
\begin{center}
  \begin{tabular}{ r || c | c }
    %\hline
    Subject \# & Dist. traversed (m) & Avg. dist. to human (m) \\ \hline 
    1 & 171.4  & 1.1\\ \hline
    2 & 161.8  & 1.13\\ \hline
    3 & 160.4  & 1.14\\ \hline
    4 & 169.9  & 1.25\\ \hline
    5 & 174.2  & 1.04\\ \hline
    6 & 166.2  & 1.3\\ \hline
    7 & 171.5  & 1.2\\ %\hline
  \end{tabular}
\end{center}
\caption {Performance of the person follower on seven subjects. Each row shows a run where the robot followed the subject on a course. The distance traversed per run is given in the second column. The average distance between the robot and the human was provided in the third column. The average distance to the person across runs are relatively consistent, and slightly higher than 1 meter because the robot got higher rewards by keeping that much of distance to the human.}
\label{tab:perf}
\end{table}

\begin{figure}[ht!]
\centering
\includegraphics[width=0.48\textwidth]{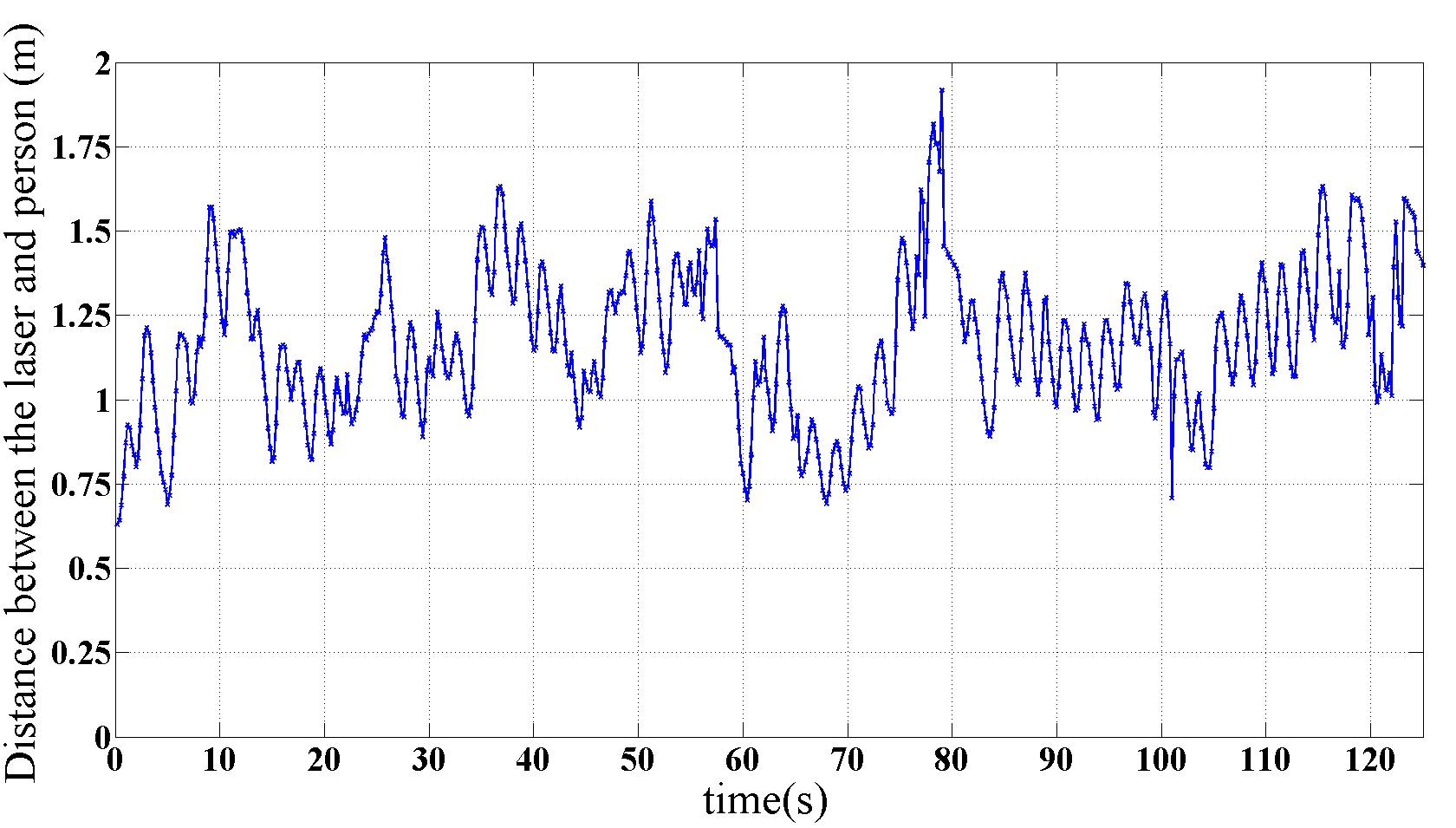}
\caption{Following distance as a function of time for one of the person following experiments.}
\label{fig:fig7}
\end{figure}

Figure \ref{fig:fig7} plots the time versus following distance for a sample run that consists of 1 lap. At $t=0$, the following is initiated and robot leaves the starting point. Around $t=8s$, the robot and the person start making a right turn. The sudden drop in following distance at $t=57.2$ signifies that the robot lost track of the person and the person tracker is reinitialized. At $t=65.6s$, the intermediate point at the end of the corridor is reached so the person makes a 180$^\circ$ turn. The robot is close to the person (about $0.75m$) around this time because the robot is rotating around itself while the person is turning back. Between $t=70$ and $t=80s$, the person walks faster than the robot, so the following distance reaches to a maximum of $1.91m$. At $t=79.8s$, the person is lost again. Around $t=115s$, the robot and the person make a left turn. The lap ends at $t=124s$. Note that the high frequency fluctuation of the following distance is a result of tracking only a single leg. There could be other contributors to this error, including inconsistent human walking speeds and noise in sensor data.

Our person following approach is applied to a telepresence robot where there is a remote user connected to the robot, as seen in Figure \ref{fig:fig6a}. In a study with then subjects, the motions of the robot was found natural, evidenced by getting an average score of 5.4 on 7-point Likert scale. The reader is referred to \cite{cosgun2013autonomous} for more details on the user study.

\subsection{Evaluation of pointing gestures}
\label{sec:eval_pointing_gestures}

To evaluate the accuracy of pointing gesture target detection, we first find the error statistics for pointing gestures and then apply it to a scenario for distinguishing two objects with varying separation.

\begin{figure}[thpb]
\centering
\includegraphics[width=80mm]{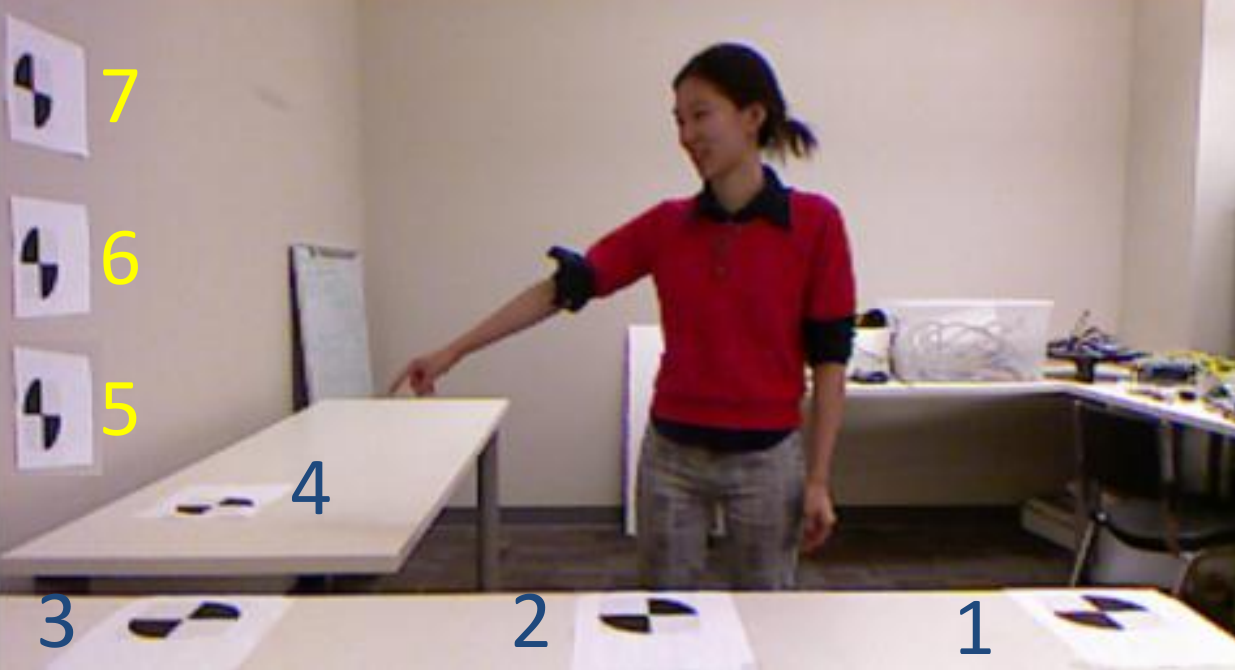}
\caption{Our study involved six users that pointed to seven targets being recorded using 30 frames per target. Four targets were placed horizontally on the table (indicated in blue) and three targets were place vertically on the wall (indicated in yellow).}
\label{fig:fig8}
\end{figure}

We collected data from six users with seven targets where people pointed at each target with their right arms (Figure \ref{fig:fig8}). Our use case is on a mobile robot platform capable of positioning itself relative to the user (Figure \ref{fig:fig6}).  For this reason, we can assume that the user is always centered in the image as the robot can easily rotate to face the user and can position itself at a desired distance from the user. The ground truth points that are represented in the camera frame are found by first finding the pixel values of targets using a corner detector, extending a virtual ray from the camera's origin to a target, and finding the intersection point to the supporting plane which is extracted from the point cloud data. We computed the mean and standard deviations of the angular errors in the spherical coordinate system for each pointing gesture method and target.

\begin{table*}[ht!]
\centering
  \begin{tabular}{c|c|c|c|c|c|c|c|c}
    %\hline
    \multirow{3}{*}{} &
      \multicolumn{4}{c|}{Target 2} &
      \multicolumn{4}{c}{All Targets} \\
    \cline{2-9}
    & \multicolumn{2}{c|}{$\theta$} & \multicolumn{2}{c|}{$\psi$} & \multicolumn{2}{c|}{$\theta$} & \multicolumn{2}{c}{$\psi$} \\
    \cline{2-9}
    & $\mu$ & $\sigma$ & $\mu$ & $\sigma$ & $\mu$ & $\sigma$ & $\mu$ & $\sigma$\\
    \hline
    Elbow-Hand & -3.8 & 6.6  & 11.3 & 10.9  & -11.2 & 7.6  & 9.6 & 6.3 \\
    \hline
    Head-Hand & 10.2  & 6.7  & -5.7 & 8.0   & -2.4 & 9.6  & -5.3 & 6.4 \\
    %\hline
  \end{tabular}
  \caption{$\mu$ and $\sigma$ of angular errors (in degrees) are given for Target 2 and across all targets. Error statistics of Table 2 was used for the object separation evaluation. The reader is referred to \cite{cosgun2015did} for the complete table.}
\label{table:pointing_error_stats}
\end{table*}

The error statistics for Target 2 and across all targets are given in Table \ref{table:pointing_error_stats}. The reason for reporting Target 2 only is that we use the error statistics of that target for the object separation study. From the data, we can tell that for the elbow-hand pointing method users typically point about $11^\circ$ to the left of the intended target direction, and about $9^\circ$ above the target direction. Similarly, the data from the head-hand pointing method reports that users typically point about $2^\circ$ to the left of the intended pointing direction, but with a higher standard deviation than the elbow-hand method.  On average, the vertical angle $\psi$ was about $5^\circ$ below the intended direction with a higher standard deviation than the elbow-hand method. The horizontal angle $\theta$ has a higher variation than the vertical angle $\psi$.  Examining the errors for individual target locations shows that this error changes significantly with the target location. Therefore, for a given target location, we first choose the closest target category in our data set, and use the corresponding mean and standard deviation values.

\begin{figure}[t!]
\centering
\includegraphics[width=0.98\columnwidth]{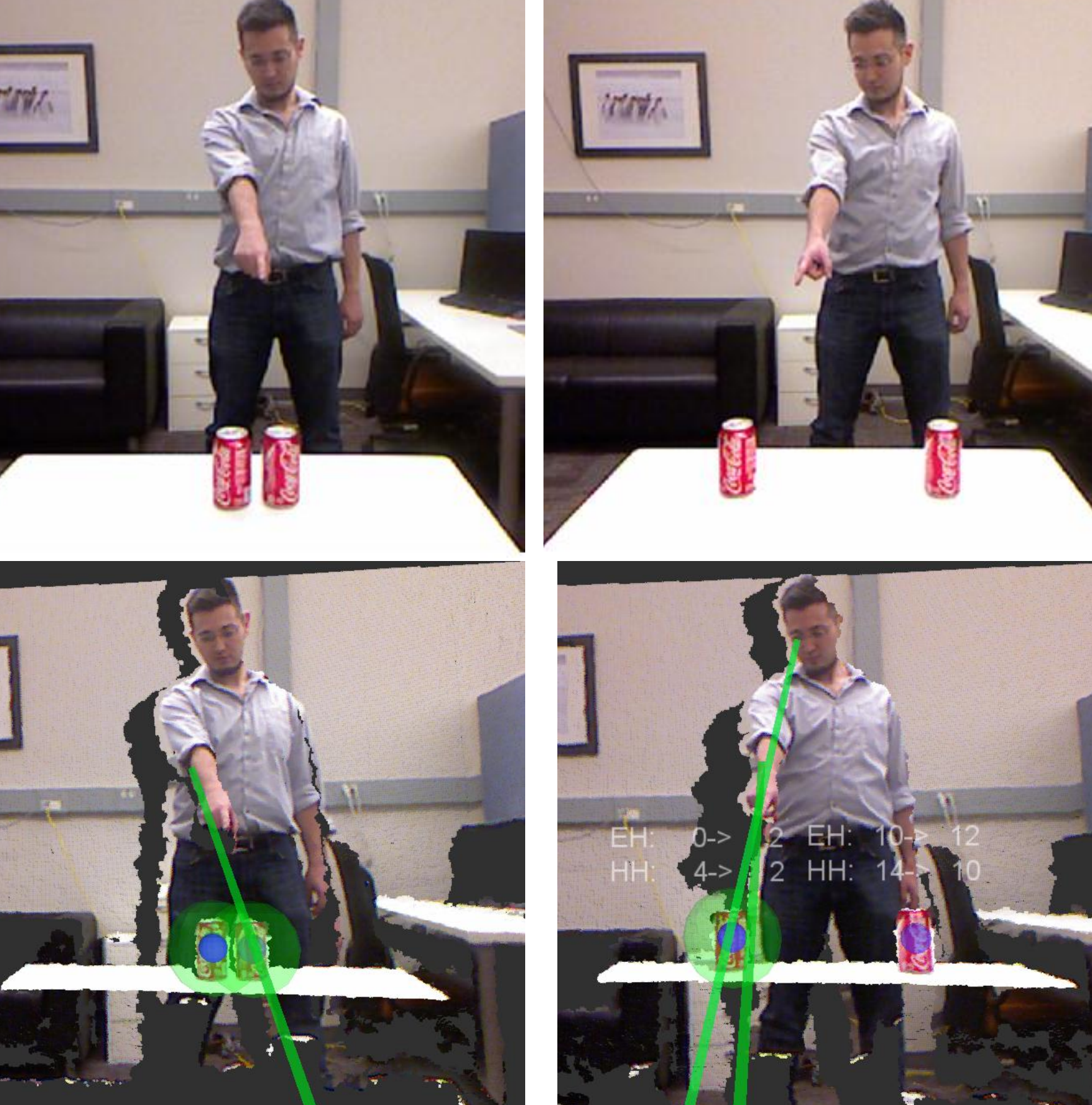}
\caption{Example scenarios from the object separation test are shown. Our experiments covered separations between $2cm$ (left images) and $30cm$ (right images). The object is comfortably distinguished for the $30cm$ case, whereas the intended target is ambiguous when the targets are $2cm$ apart. Second row shows the point cloud from the RGB-D camera's view. Green lines show the Elbow-Hand and Head-Hand directions whereas green circles show the objects that are within the threshold $D_{mah}<2$.}
\label{fig:fig9}
\end{figure}

Next, we conducted an experiment to determine how our approach distinguished two potentially ambiguous pointing target objects. The setup consisted of a table between the robot and the person and two coke cans on the table (Figure~\ref{fig:fig9}) where the separation between objects was varied. The center positions of objects were calculated in real-time by a point cloud segmentation with supporting plane assumption. The separation between objects were varied with 1 cm increments from $2 cm$ to $15 cm$ and with $5 cm$ increments between $15 cm-30 cm$. We could not conduct the experiment below $2 cm$ separation because of the limitations of our perception system. The experiment was conducted with one user who was not in the training dataset. For each separation the user performed five pointing gestures to each object. The person pointed to one of the objects and the Mahalanobis distance $D_{mah}$ to the intended object and the other object is calculated. Error statistics of Target 2 (Figure \ref{fig:fig8}) was used for this experiment.

\begin{figure}[ht!]
\centering
        \subfigure[]{%
            \label{fig:fig10a}
            \includegraphics[width=0.44\textwidth]{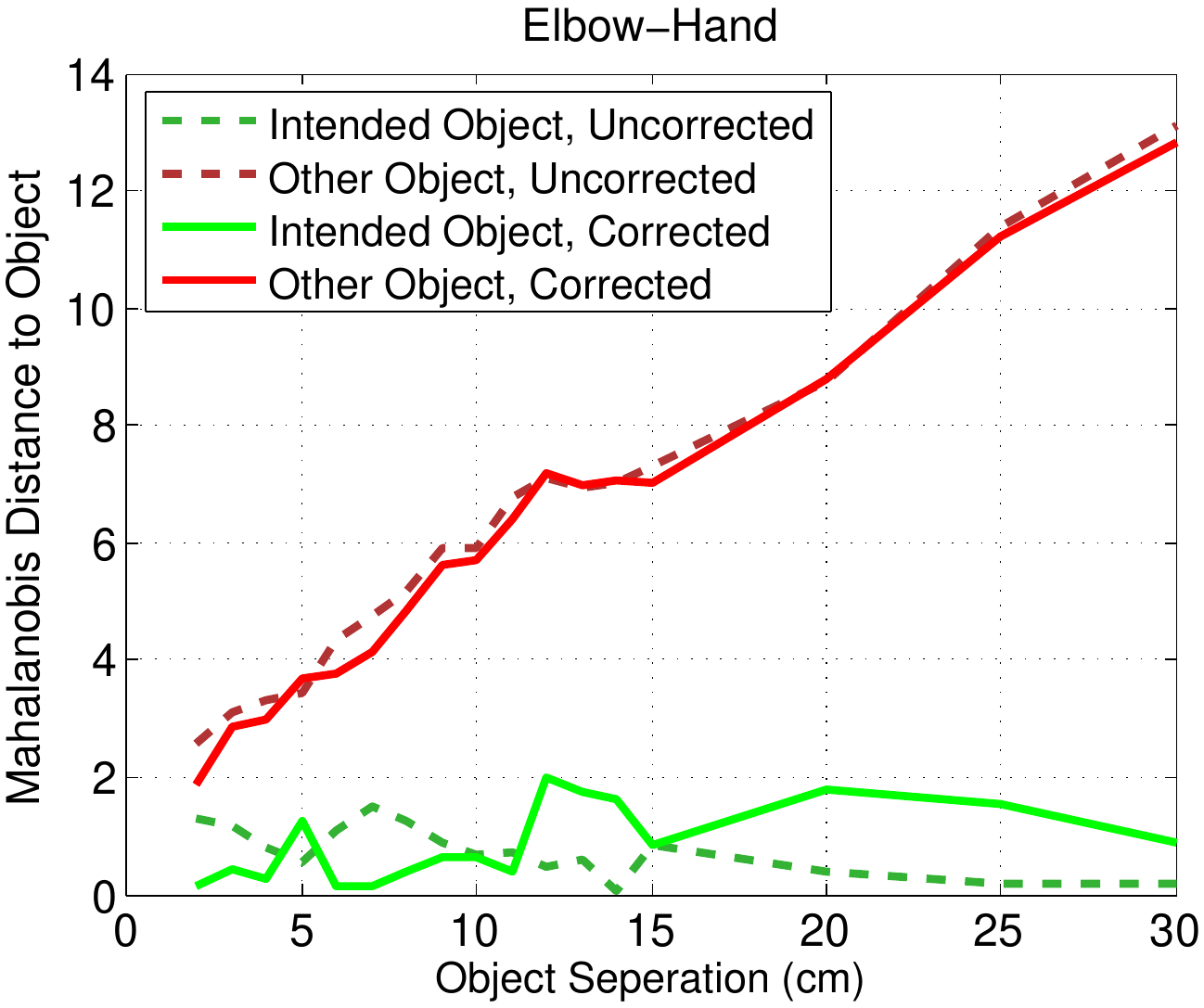}
        }%
        \\
        \subfigure[]{%
           \label{fig:fig10b}
           \includegraphics[width=0.44\textwidth]{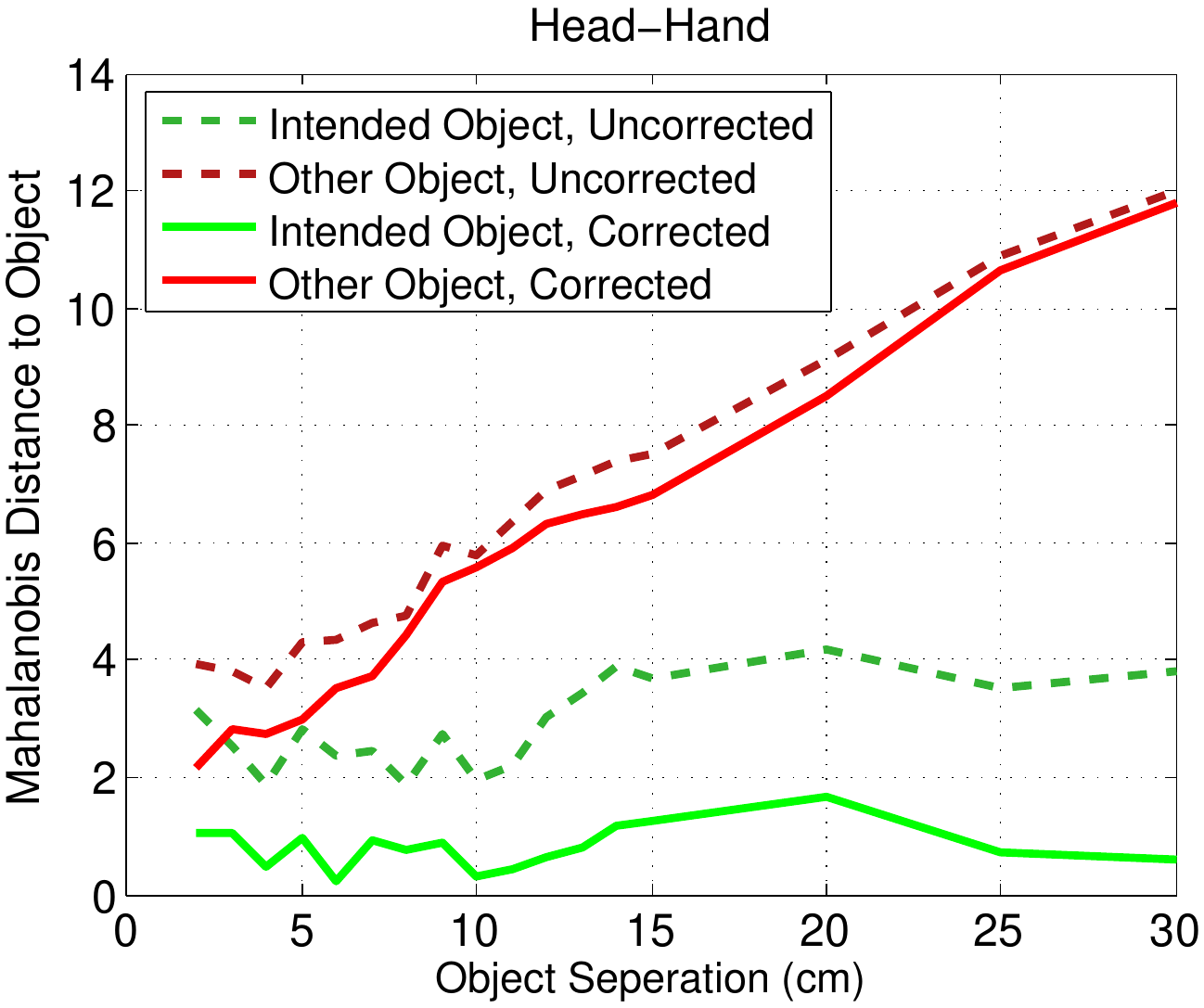}
        }        
    \caption{%
	Resulting Mahalanabis distances of pointing targets from the Object Separation Test is shown for a) Elbow-Hand and b) Head-Hand pointing methods. Plots for intended objects are shown in green and for the unintended objects are shown in red. Solid lines show distances after correction is applied. Less Mahalanobis distance for intended object is better for reducing ambiguity.
     }%
   \label{fig:fig10}
\end{figure}

The results of the object separation experiment is given for Elbow-Hand (Figure~\ref{fig:fig10a}) and Head-Hand (Figure~\ref{fig:fig10b}) methods. The graphs plot object separation versus the Mahalanobis distance for the intended unintended objects for corrected and uncorrected pointing gestures. First, the Mahalanobis distance $D_{mah}$ for the intended object was always lower than the other object. The corrected $D_{mah}$ for both Elbow-Hand and Head-Hand methods for the intended object was always below 2. Because of that, we chose the threshold $D_{thres}=2$. We notice that some distances for the unintended object at 2cm separation is also below $D_{mah}<2$. Therefore, when the objects are 2 cm apart the pointing target becomes ambiguous for this setup. For separations of 3cm or more, $D_{mah}$ of the unintended object is always over the threshold so there is no ambiguity. Second, correction significantly improved Head-Hand accuracy at all separations, slightly improved Elbow-Hand between 2-12cm but slightly worsened Elbow-Hand after 12cm. Third, the Mahalanobis distance stayed generally constant for the intended object, which was expected. It linearly increased with separation distance for the other object. Fourth, patterns for both methods are fairly similar to each other, other than Head-Hand uncorrected distances being higher than Elbow-Hand.

\section{Robot navigation using semantic maps}
\label{sec:robot_navigation}

Autonomous navigation is one of the most fundamental capabilities for a mobile robot. There are many approaches that achieve point-to-point autonomous navigation thanks to the advances in mapping, localization and motion planning research. Many of these algorithms are optimized to find the least-cost path or the shortest path. However, often there are additional social factors to consider for navigation among humans. 

First, it is not natural for humans to provide the goals in exact coordinates. Instead, the robot should be able to understand goals that are expressed in natural language and grounded with shared references. In our approach users provide annotated landmarks as goals to the robot using a mobile app. Our method of goal calculation from a user query is discussed in Section \ref{sec:finding_the_goal_point}.

Second, robots should pay special attention to their motions when there are humans in the environment, or when there is a probability of encountering a human. For example, while it is acceptable for a robot to get very close to a wall, doing so to a human is socially unacceptable and unsafe. Similarly the sudden appearance of a robot can surprise humans and cause discomfort. There are many other social scenarios where the shortest path may not be optimal. Therefore, context-aware path planning algorithms should treat humans and obstacles differently to enable intelligent robot behaviors. Our human-aware path planning method is described in Section \ref{sec:human_aware_path_planning}.

Third, semantic maps could be exploited to enhance the navigation behaviors. Robot navigation behaviors could be tailored to the task at hand. For example, when the robot is following a person during the map labeling task the robot chooses its sub-goals to facilitate interaction. Similarly, semantic maps could be useful to negotiate passing in bottlenecks, such as door passages. We present our context-aware person following approach in Section \ref{sec:context_aware_person_following} built on top of the person following method presented in Section \ref{sec:person_following}.

\subsection{Navigating to labeled landmarks}
\label{sec:navigating_to_labeled_landmarks}

\subsubsection{Finding the goal point}
\label{sec:finding_the_goal_point}

In our current application, the goals are either labeled landmarks or objects using a phone app. When the user enters a landmark as a navigation goal, the robot first finds a goal point in the metric map, then plans and executes a socially acceptable path toward this goal.

\begin{figure}[t!]
\centering
        \subfigure[]{%
            \label{fig:fig11a}
            \includegraphics[width=0.48\textwidth]{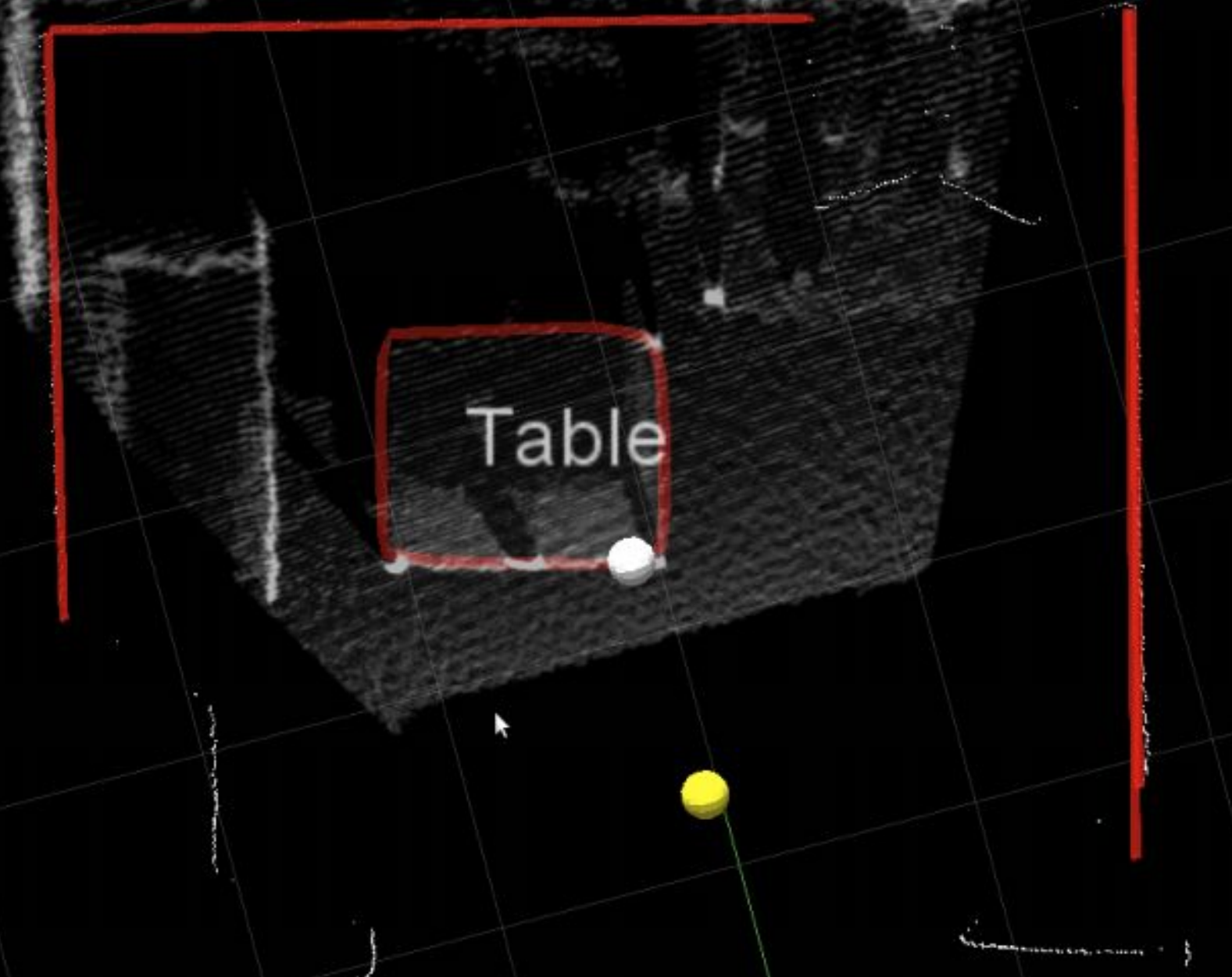}
        }%
        \\
        \subfigure[]{%
           \label{fig:fig11b}
           \includegraphics[width=0.48\textwidth]{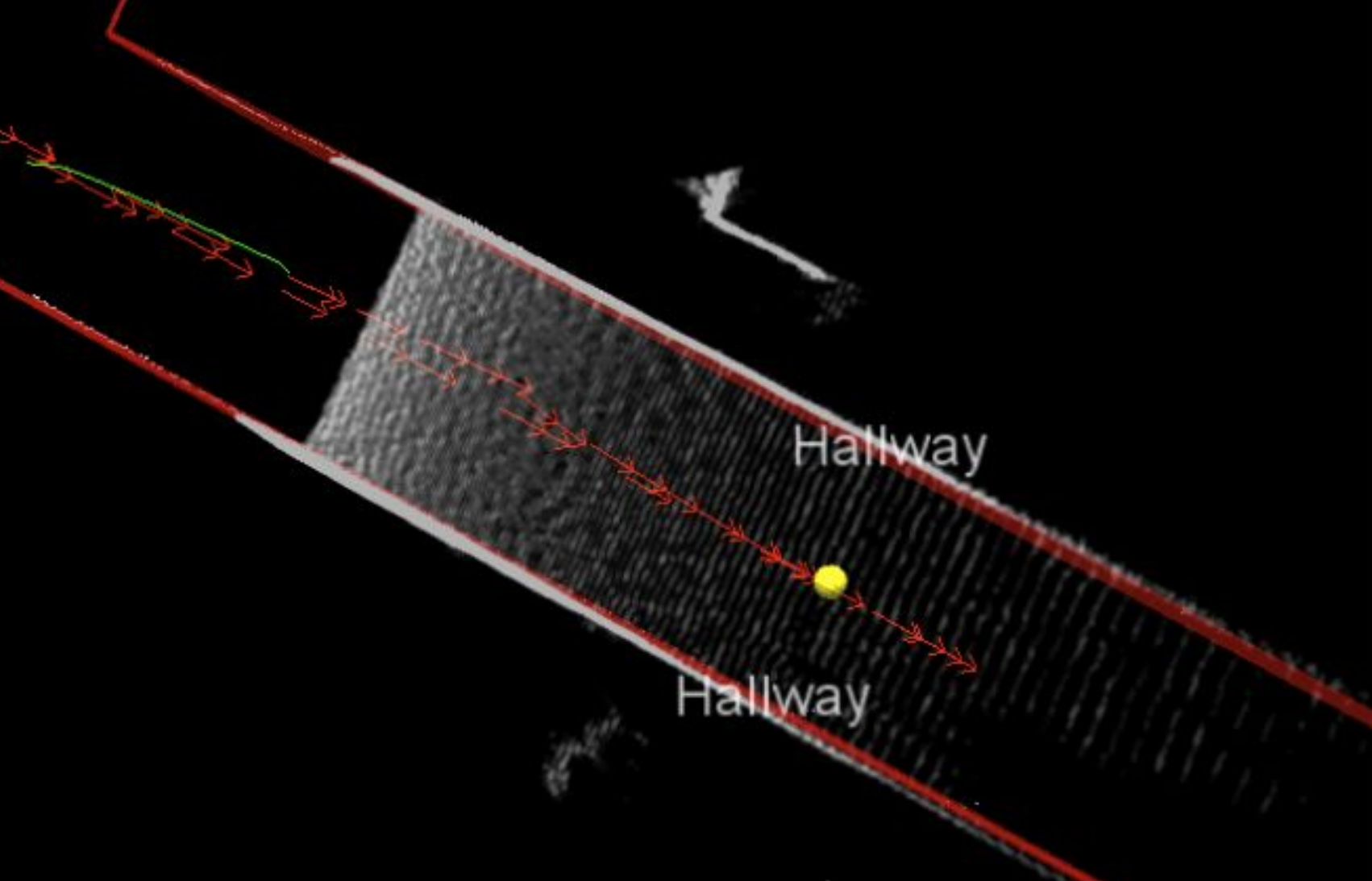}
        } 
\caption{a) Top down point cloud view of a room. A planar landmark with the label $Table$ has previously been annotated by a user. The convex hull for the planar landmark is shown in red lines. When asked to navigate to $Table$, the robot calculates a goal position, which is shown as the yellow point; b) Top down point cloud view of a hallway. The user has previously annotated two planar landmarks with the same label: $Hallway$. When asked to navigate to $Hallway$, the robot chooses a goal position in the middle of the planar landmarks which is shown as the yellow point.}
\label{fig:fig11}
\end{figure}

When a planar landmark is entered as the goal, the metric goal is chosen towards the closest edge of the plane. We first select the closest vertex on the landmark's convex hull to the robot's current position, and projects it to the floor plane. We find the line between the closest vertex on the convex hull and the robot's current pose. The goal position is selected to be on this line, a meter away from the vertex. With this design, the robot would be in close proximity of the desired planar surface and be oriented towards it. This method is suitable for horizontal planes, such as tables, and vertical planes alike, such as doors. An example for finding a goal pose for a uniquely labeled planar landmark is shown in Figure \ref{fig:fig11a}.

When there are multiple planes associated with the same label we interpret this landmark as a region or space, such as a room or corridor. In this case, we project the points of all planes with this label to the ground plane and compute the convex hull. The goal position is chosen as the centroid of the convex hull and the goal orientation is unspecified, meaning the robot would not change its orientation upon reaching the goal position. An example goal position where the goal landmark label ``hallway'' represents two walls enclosing a hallway is shown in Figure \ref{fig:fig11b}.

\subsubsection{Human-aware path planning}
\label{sec:human_aware_path_planning}

After a goal position is calculated, a socially acceptable path is planned starting from the current position of the robot. Most approaches divide the robot path planning problem into two: global and local planning. Our approach adopts this template and further divides global planning into two parts: static and dynamic planners. The static planner finds a path on the map of the environment by considering the safety and disturbance of humans, as well as the path length but does not consider the future movements of humans. The dynamic planner simulates the future motions of humans by using a social motion model and refines the static path. The dynamic planner takes the static path and refines it by considering the predicted temporary goals of humans and their reaction to the robot's future movements. The predicted goals are used to forward-simulate the human trajectories and generate `social forces' for the social motion model. A part of the robot's path is then recomputed in compliance with the model.

\begin{figure}[t!]
\centering
\includegraphics[width=0.48\textwidth]{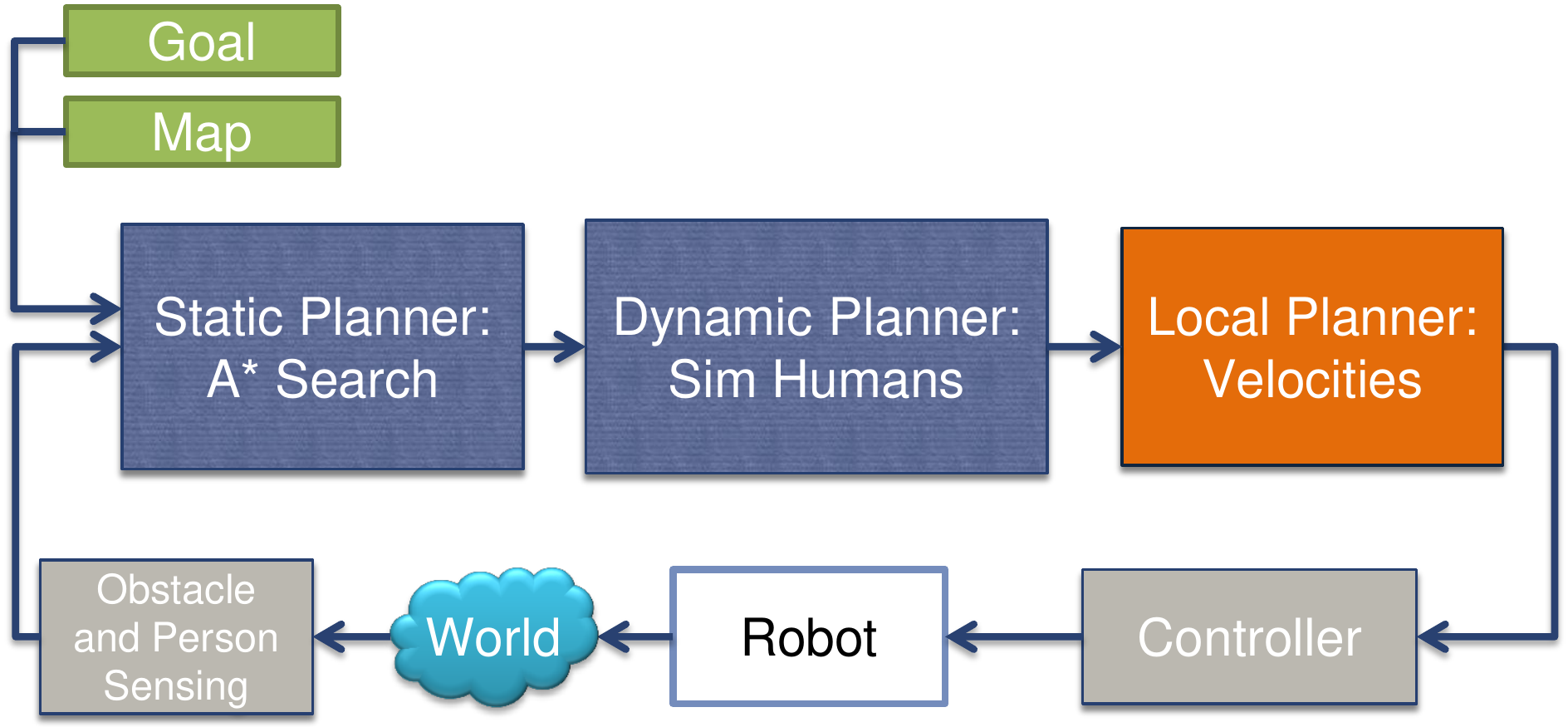}
\caption{System overview. When a map and goal is provided to the static planner obstacles and humans are detected and then a path is planned using A* search. The dynamic planner refines the static plan by simulating human reactions to the robot motion. The local planner receives the result, and computes the linear and angular velocities necessary to follow the path. The controller applies these velocities to the robot, which in turn acts in the world. The sensors generate new data, and the loop restarts.}
\label{fig:fig12}
\end{figure}

Our local planner is a trajectory planner that computes the linear and angular velocities that would allow the robot to follow the dynamic path. The navigation system overview is shown in Figure~\ref{fig:fig12}. The obstacles are differentiated from humans in two modules: 
\begin{enumerate}
\item Static planner: Approaching humans add a safety cost and traversing between humans add a disturbance cost.
\item Dynamic planner: Future motion of the humans are simulated, which is then used to refine the static path.
\end{enumerate}
More information about this work can be found in \cite{cosgun2016anticipatory}.

The static planner takes the start and goal positions and a 2D grid map as input and aims to find a set of waypoints that connects the start and goal cells. The output path has the minimum cost using a linearly weighted cost function with three components: path length, safety and disturbance. We use A* search with Euclidean heuristics on a 8-connected grid map to find the minimum cost path. The path length cost is the total length of a path. The safety cost aims to model personal spaces of people. A gaussian cost function is attached to each person in the environment. The safety cost of a cell is the maximum safety cost value among all humans in the environment. The disturbance cost aims to represent the cases where the robot potentially disturbs the interaction of a group of humans. For example, if two people are facing each other and talking, then the robot should not cross between them. The disturbance cost is a non-zero cost if the robot's path crosses between two people who are in reasonable proximity to each other. We do not detect if there actually is conversation between the people but estimate the disturbance cost using body poses of agents. This cost increases if the body orientations of two people are facing each other and is inversely proportional to the distance between a pair of humans. Detecting human formations is a challenging task that has been addressed in the literature \cite{cristani2011social}, however we compute this cost for each pair of humans in the scene without explicitly detecting the formation. This works because the disturbance cost becomes zero after a cut-off distance threshold.

The dynamic path refinement processes the static plan by simulating parts of the path where group of humans are closeby. We use the Social Forces Model (SFM) \cite{helbing1995social} to simulate the motions of humans and the robot. Interactions between people are modeled as attractive and repulsive forces in SFM, similar to potential fields. The forces are recomputed iteratively and the resulting simulated path sections replaces the corresponding path sections in the static plan. We use DWA as the local trajectory planner. The refinement step allows considering future motions of humans due to the robot's future motions.

\begin{figure}[ht!]
\centering
        \subfigure[]{%
            \label{fig:fig13a}
            \includegraphics[width=0.97\columnwidth]{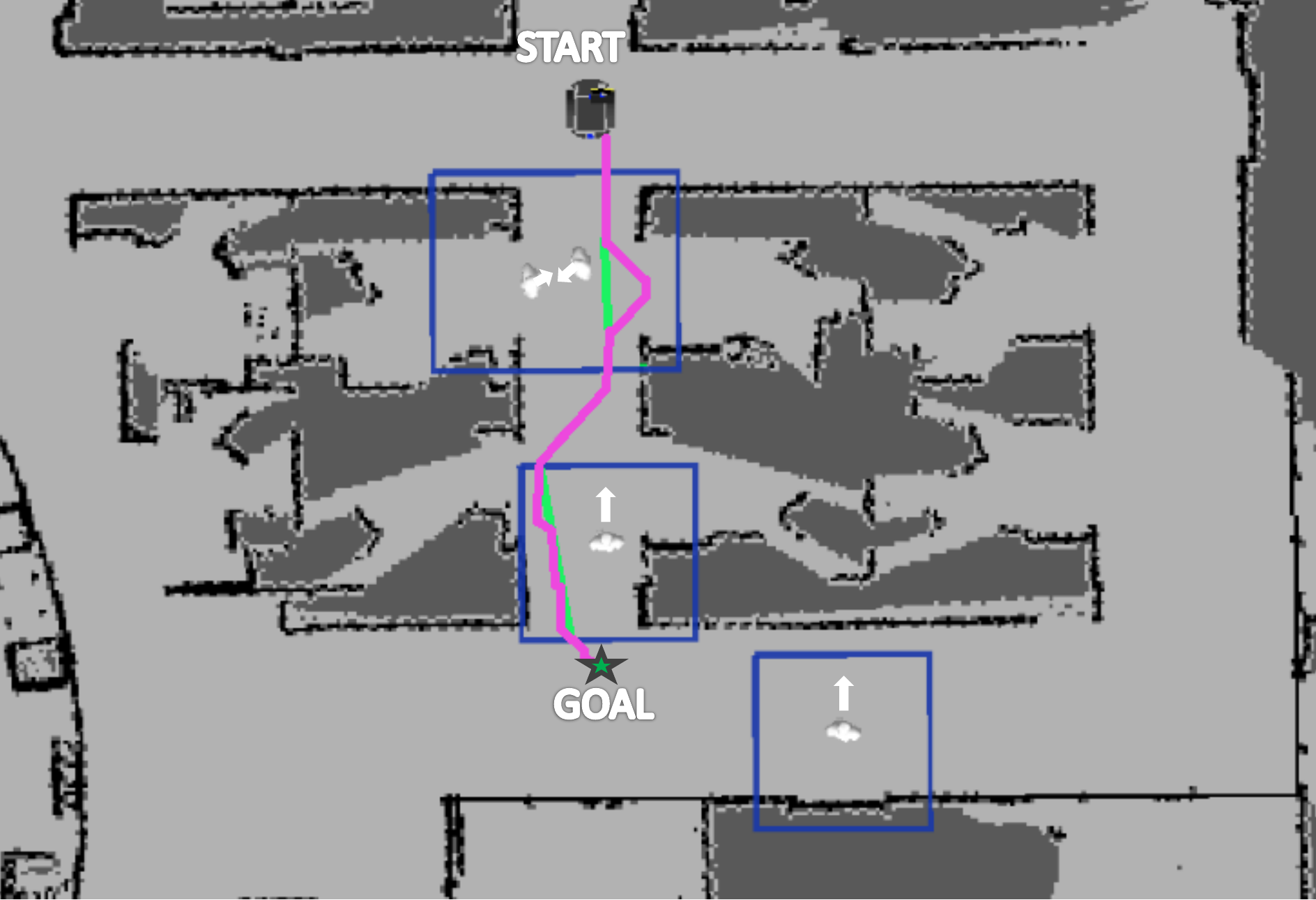}
        } 
        \subfigure[]{%
           \label{fig:fig13b}
           \includegraphics[width=0.97\columnwidth]{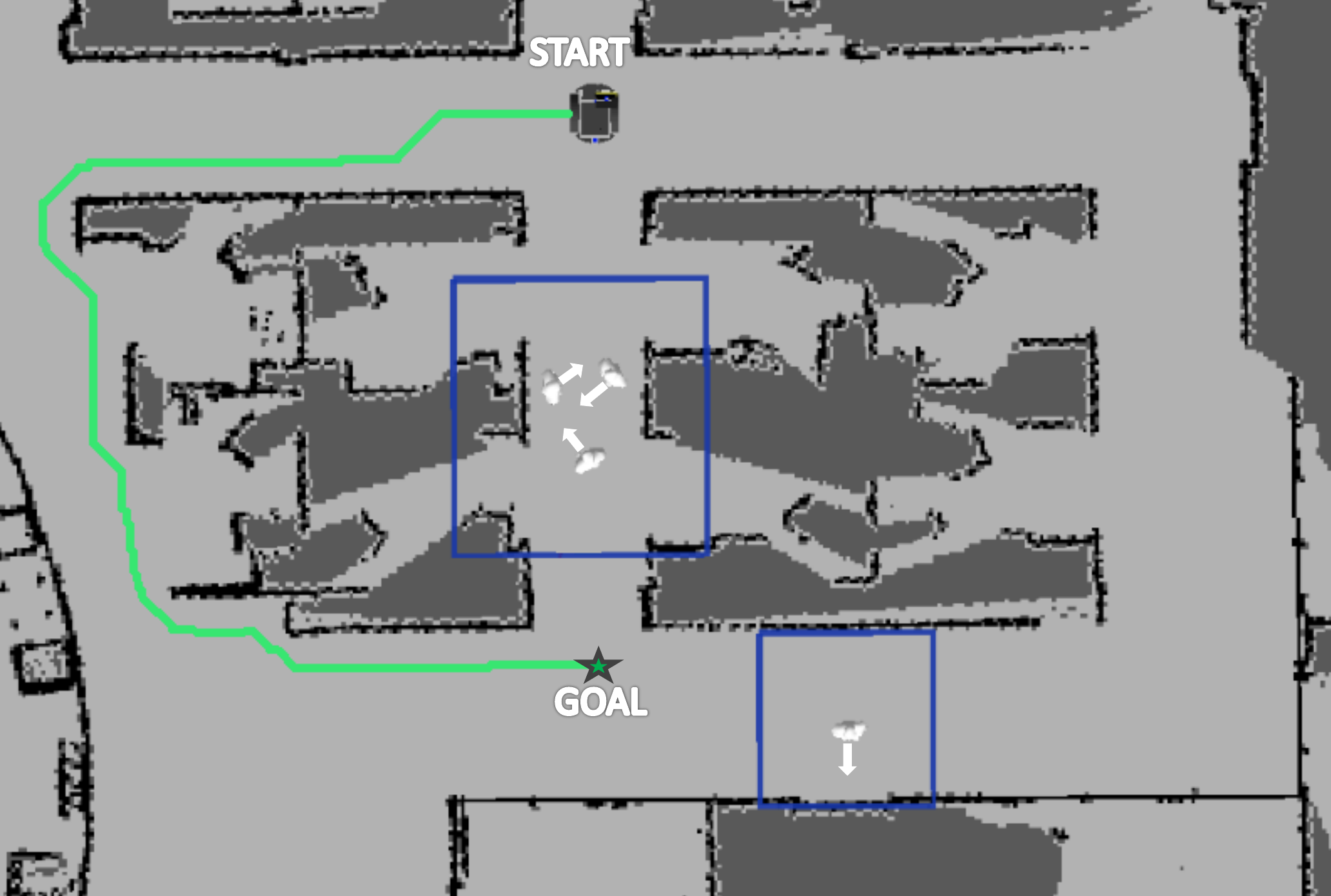}
        } \\
	\subfigure[]{%
           \label{fig:fig13c}
           \includegraphics[width=0.97\columnwidth]{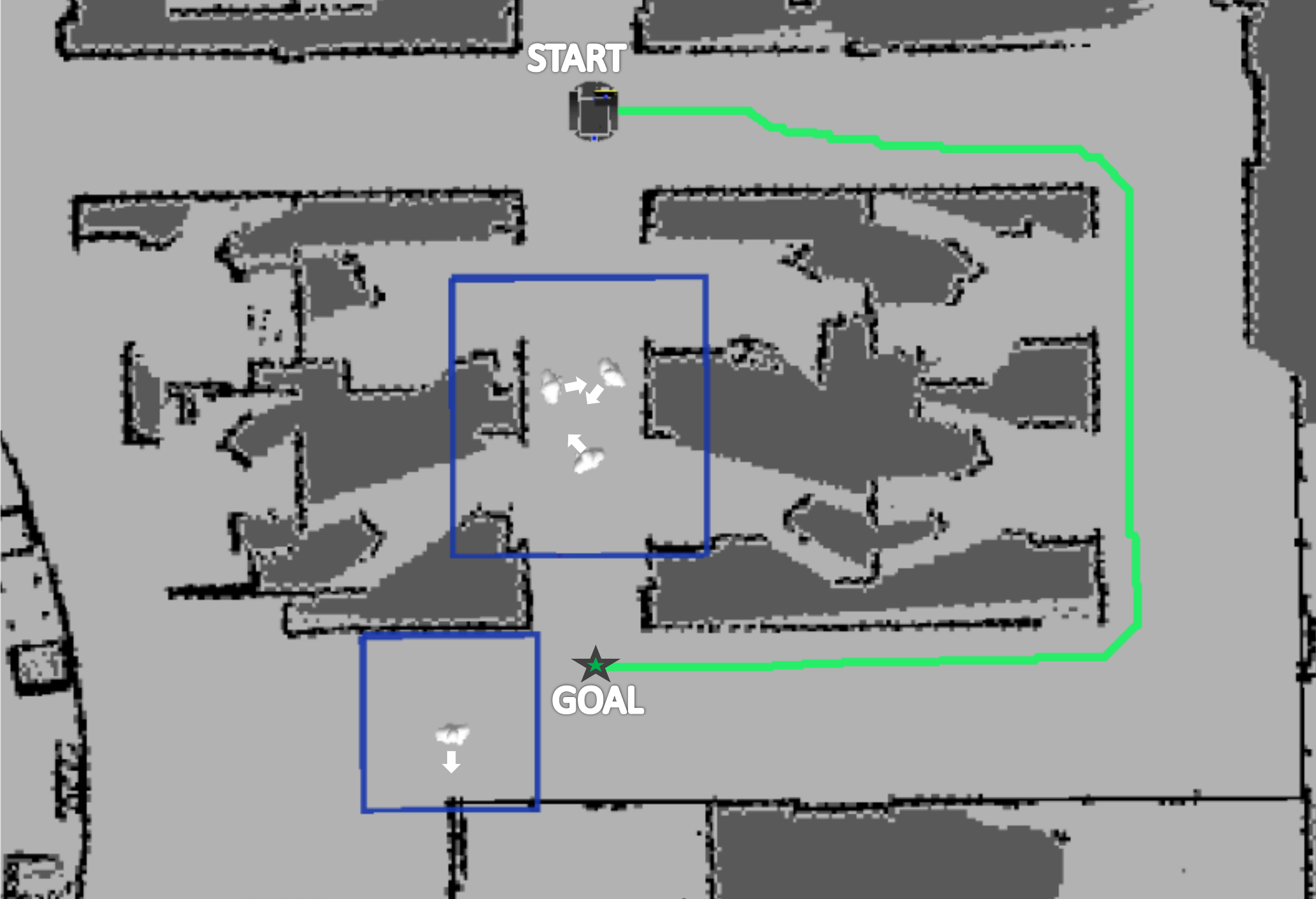}
        }
        
    \caption{%
	Path planner's output differ given the poses and grouping of humans. a) The robot takes shortest route, traveling in the vicinity of a group of two and another individual; b) third individual joins the group. Robot takes a longer path that doesn't have humans on path; c) fourth person changes his position, leading the robot to take the longest route.
     }%
   \label{fig:fig13}
\end{figure}

We demonstrate our approach with an example in simulation (Figure~\ref{fig:fig13}). The goal of the robot is to navigate to a goal position in an office environment where there are four people present in the environment. In this scenario, we show how the path changes significantly when only poses of humans are varied. There are three main ways the robot can navigate to its goal: left, center or right corridor.

In the first configuration in Figure~\ref{fig:fig13a}, two people are grouped together as they are looking at each other and likely conversing. The robot decides to take the center corridor. First, it slightly disturbs the speaking duo, then switches sides in the corridor, and reaches its goal. In the figure, the dynamic path (pink line) is overlaid on the static path (green line). 

In the second configuration in Figure~\ref{fig:fig13b}, the third person at the center corridor joins the conversation. Now we have two group regions (rectangles) in the scene. Since passing through a group of three people would introduce a high disturbance cost in addition to the safety cost, the robot decides to take a longer route (left corridor). Since this path does not intersect any group regions the dynamic simulation was not conducted.

In the third configuration in Figure~\ref{fig:fig13c}, the group of three hasn't moved, but the fourth person has changed its position. In this case, if the left corridor is taken again, an additional safety cost would be incurred. Therefore the robot decides to take the longest route (right corridor). Again, since the robot travels far from humans the dynamic simulation was not conducted.

\subsection{Context aware person following}
\label{sec:context_aware_person_following}

As briefly reviewed in Section \ref{sec:rel_context_aware_navigation}, most person following methods in the literature have the same underlying principle: a target position is calculated given the human's position each iteration and a control method finds actions iteratively to navigate towards that position. This results in reactive robot behaviors where the robot follows the human blindly irrespective of the task and context. Our person following method presented in Section \ref{sec:person_following} also falls under this category.

Although reactive methods are sufficient for some scenarios, it can easily lead to deadlock scenarios. For example, consider the case that the followed person goes through a door and stops just outside the doorway. In this case, the robot would occupy the doorway, blocking other people's passage, however does not know it caused an undesirable social situation. If the robot knows what the user intends to do, it can anticipate those actions and suitably adjust its behavior. Person following can be used in different contexts, such as for carrying luggage in airports or groceries in a supermarket. We showed in previous sections that semantic information could be used to communicate goals between the robot and the user. The stored semantic information can also be used to facilitate robot navigation.

We model the task scenario during person following as a state machine, where transitions are triggered via events. A general scenario during person following is implemented as a sequence of four phases:

\begin{enumerate}
\item Signal: The robot detects an event using perceptual cues.
\item Approach: The robot moves to a position better suited to the task.
\item Execution: The robot and/or the human executes the task.
\item Release: The robot detects the end of event and continues with the basic following behavior.
\end{enumerate}

We focus on two specific scenarios of context-aware person following using the 4-phase model: following for labeling in Section \ref{sec:following_for labeling} and passing doors in Section \ref{sec:following_door_passing}. For both of the scenarios, we demonstrate the capability with a user who is knowledgeable of the robot's capabilities.

\subsubsection{Following for interactive labeling}
\label{sec:following_for labeling}

\begin{figure}[ht!]
\centering
        \subfigure[]{%           
           \label{fig:fig14a}
           \includegraphics[width=0.3915\textwidth]{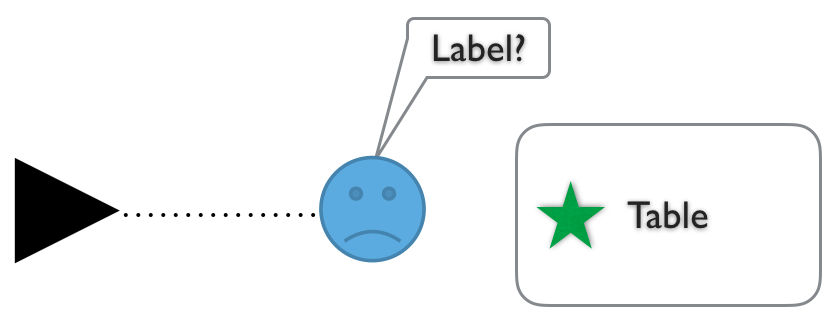}
        }
         \subfigure[]{%           
           \label{fig:fig14b}
           \includegraphics[width=0.2565\textwidth]{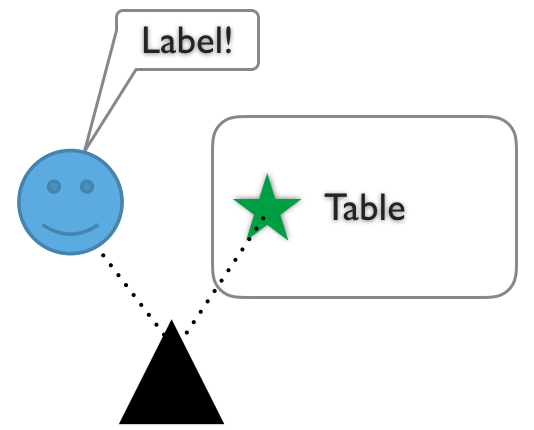}
        } 
    \caption{a) A common problem encountered during person following for interactive labeling. The user wants to label an object on the table, however, the robot does not know the user's intention and stays behind at a fixed distance to the user; b) Our solution is for the robot to navigate to a location that gives the robot a better chance to observe the user and the object simultaneously.}
   \label{fig:fig14}
\end{figure}

We first examine the person following scenario for interactive labeling of semantic landmarks as described in Section \ref{sec:interactive_map_labeling}. For this scenario the robot follows the user as he/she moves between the landmarks or objects of interests. Sometimes when a user wants to label an object, undesirable social situations can occur because the robot does not have the task context. In this example, the context is defined as the understanding of being a part of a collaborative task: interactive labeling. When a user stops in front of a landmark or object to label it, if the robot stays behind it can not perceive the pointing gesture and the landmark at the same time. This situation is illustrated in Figure \ref{fig:fig14}. 

The robot can behave more intelligently if the robot can predict ahead of time when the user is going to label a landmark. When the robot detects that the user intends to label a landmark or object our approach is to position the robot base so it has a better chance to perceive both the pointing gesture and the object/landmark of interest.

\begin{figure*}[h]
\centering
        \subfigure[]{%           
           \label{fig:fig15a}
           \includegraphics[width=0.445\textwidth]{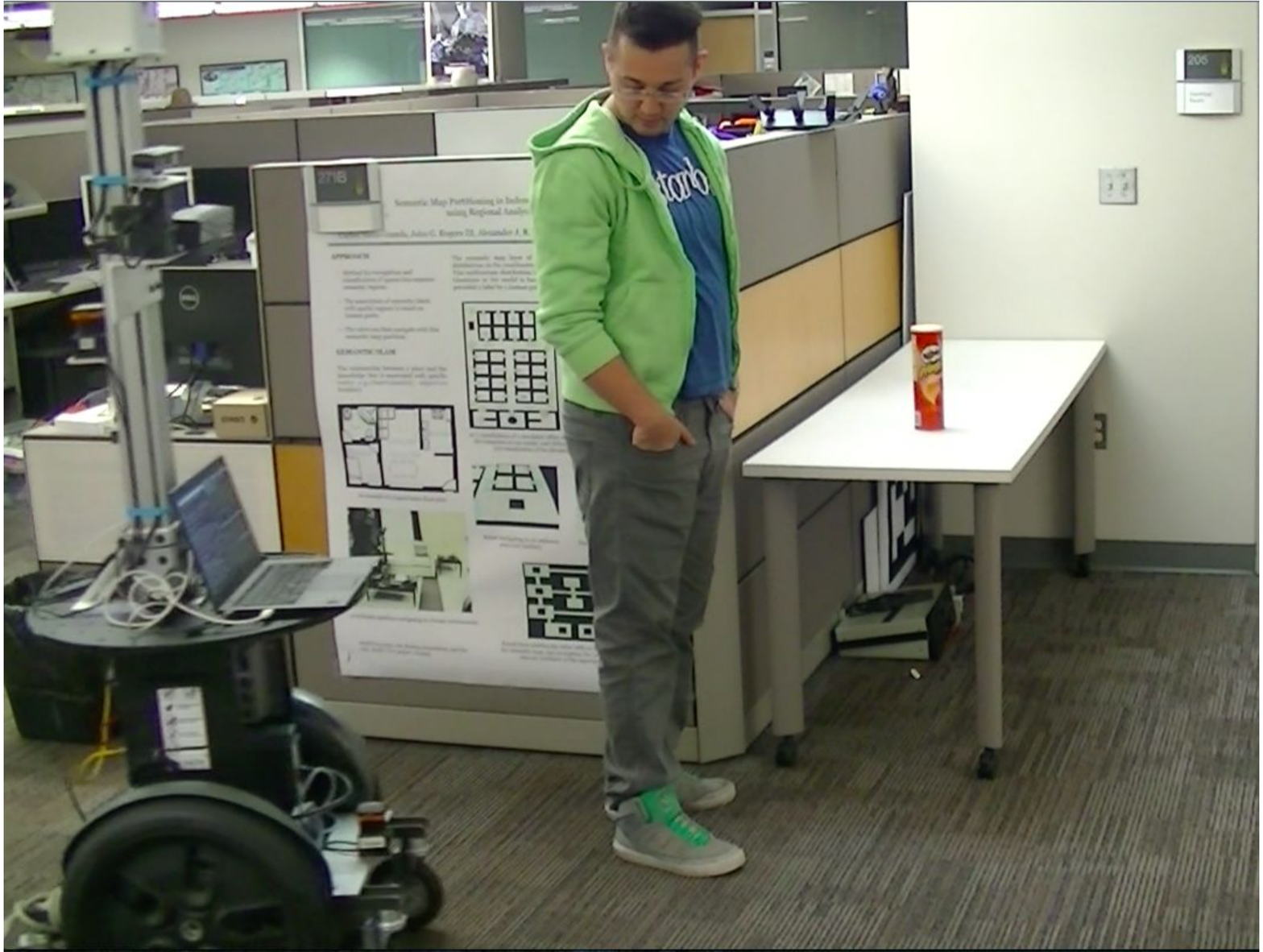}
        }
         \subfigure[]{%           
           \label{fig:fig15b}
           \includegraphics[width=0.45\textwidth]{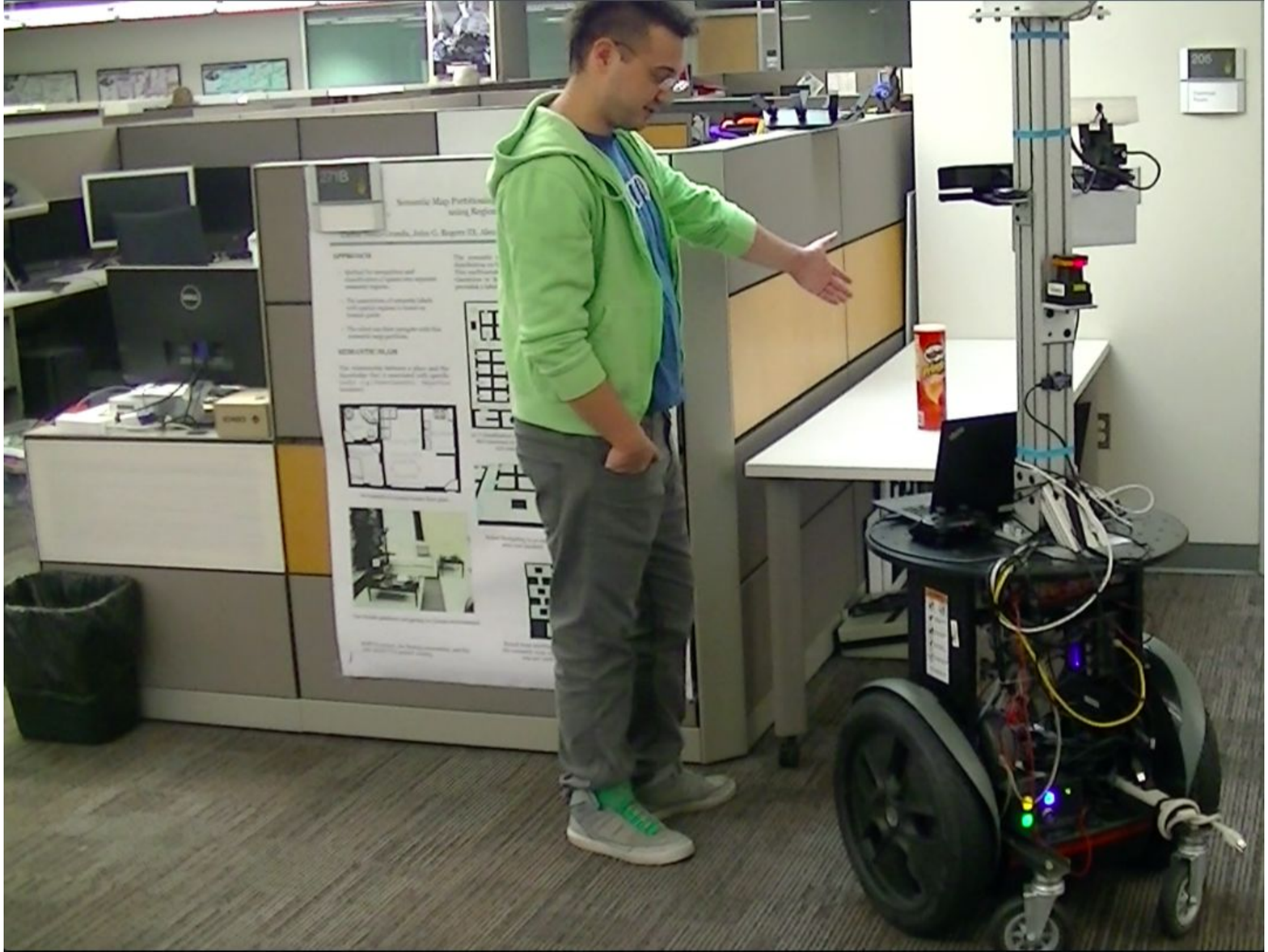}
        } \\
        \subfigure[]{%
        	\label{fig:fig15c}
            \includegraphics[width=0.45\textwidth]{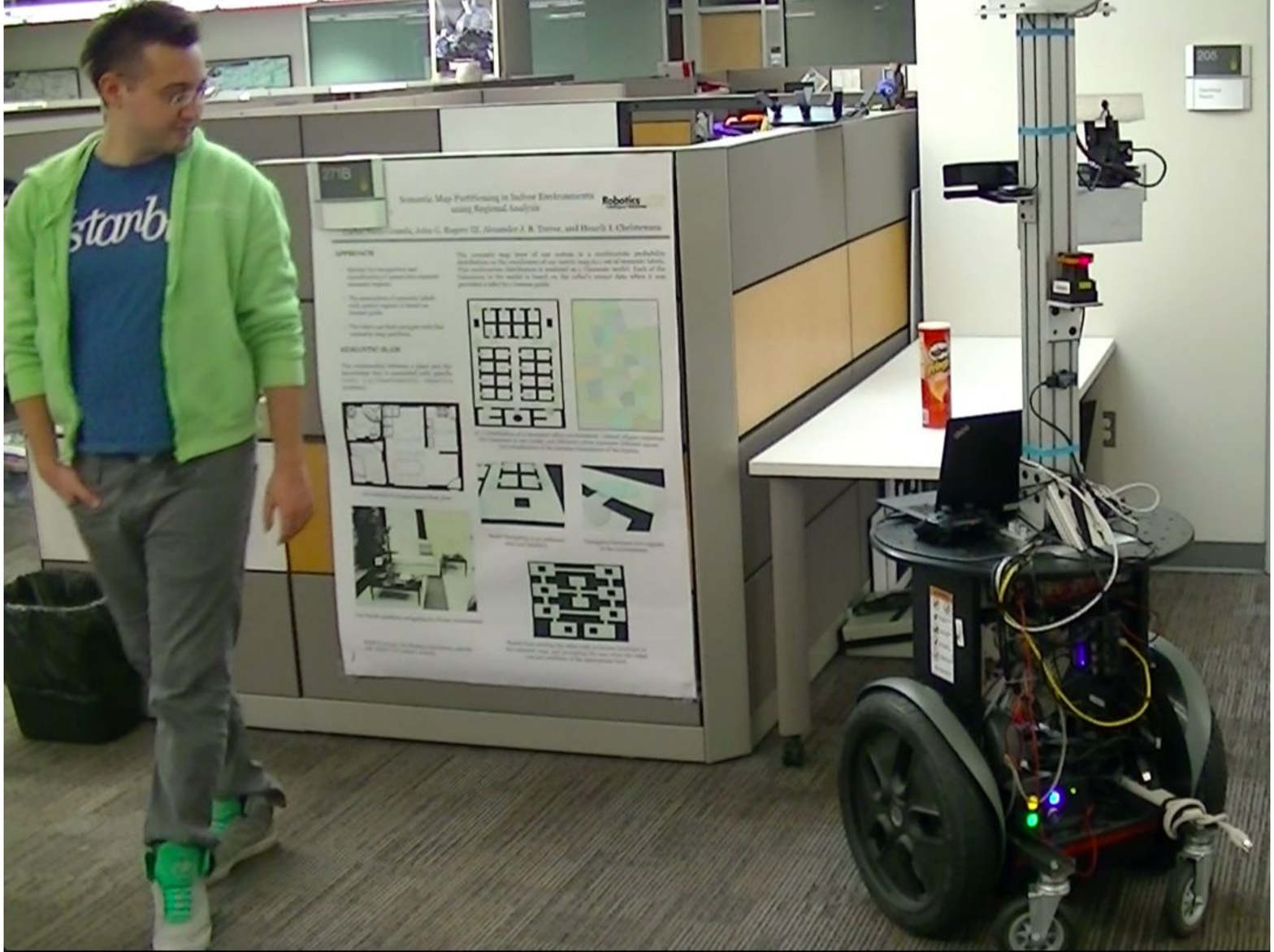}
        }%\\        
        \subfigure[]{%           
           \label{fig:fig15d}
           \includegraphics[width=0.46\textwidth]{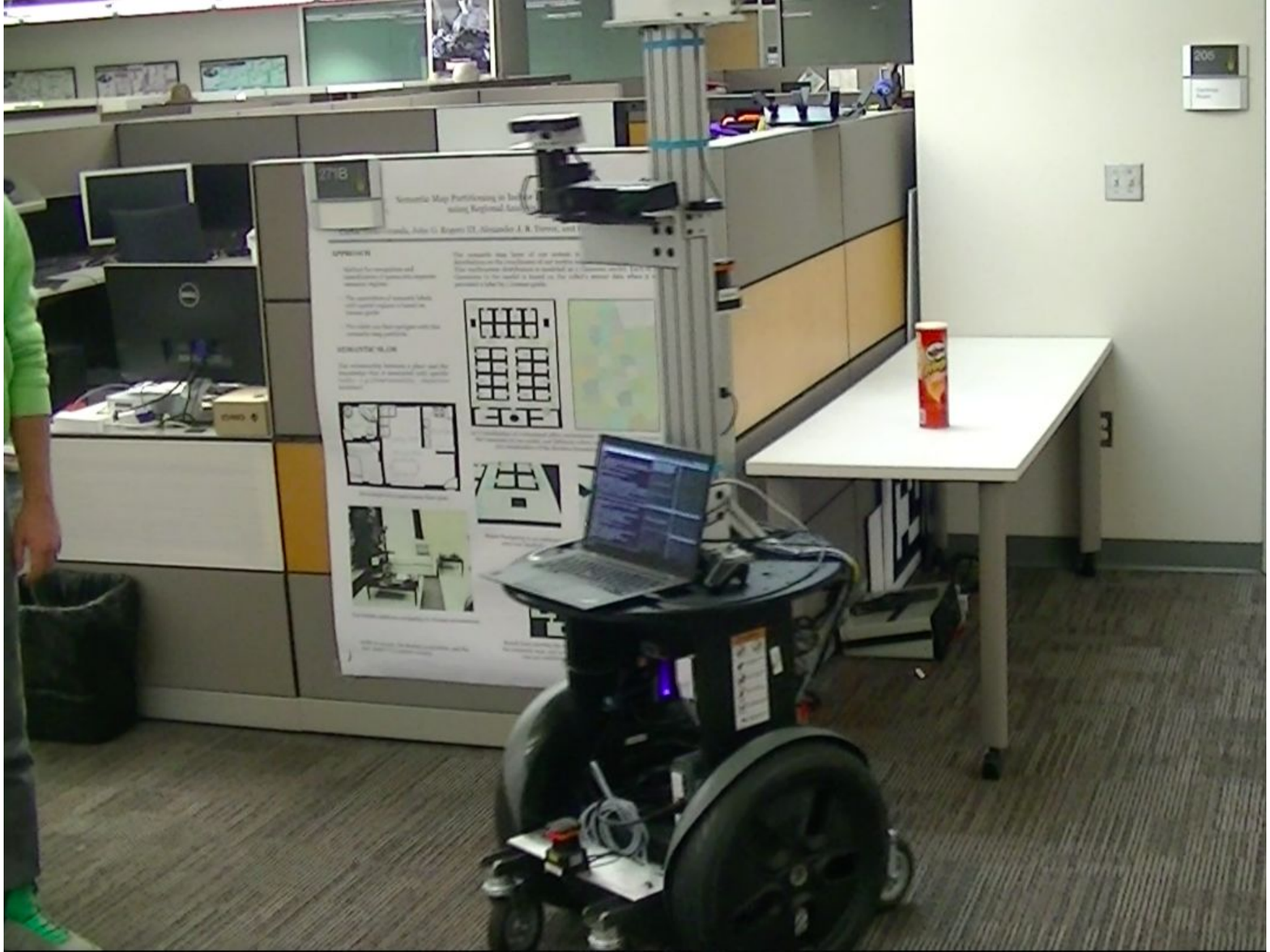}
        }
    \caption{Demonstration of context-awareness for interactive labeling. The robot is following the user throughout the environment and keeping a fixed distance of $1.2m$ to the user. a) Signal phase: The user has stopped and is in the close proximity to the convex hull of the table; b) Approach phase: The robot calculates and navigates to a goal position, so it can perceive the pointing gesture and target. Execution phase: The user points out to the object on the table; c) Release phase: the user moves away from the table d) Basic following behavior continues.}
   \label{fig:fig15}
\end{figure*}

\begin{table*}[t!]
    \centering		
  \begin{tabular}{l |  m{10cm}}
    %\toprule
    Signal & {dist(user, convex hull(landmark))$<$threshold}\\       
	                           & {speed(user)$\sim$0} \\
	                           & {person roughly facing landmark}\\ \hline		                           		                                
    Approach & {Optimal goal: Close to both the landmark and person, facing in between}\\       \hline
    Execution & {User points and labels landmark}\\  \hline
    Release & {dist(user, convex hull(landmark))$>$threshold}\\ 
    %\bottomrule
  \end{tabular}
	\caption{Conditions to trigger phases when the user is involved with the Landmark Labeling Event during following.}    
    \vspace{-0.3cm}
    \label{table:situation_aware_list_landmark}
\end{table*}

We follow the 4-phase behavior design for person following for interactive labeling. The Signaling phase is triggered whenever the user is close to a unlabeled landmark in the semantic map. The user must have close to zero speed to enable signaling for this behavior, because the user may walk past the landmark. After the robot detects the signal we sample positions around the group to locate a ``suitable'' goal pose for the robot. A pose that is collision free but that gives the robot highest chance of interaction is favored. A suitable goal position should be at an equal distance to the landmark and the user, and the goal orientation should be selected so the robot faces in between the landmark and the user. Moreover, the goal point should not be very close to an obstacle. We linearly sample points around the ``group'' formed by the user and the landmark's centroid. The points are sampled from the \textit{p-space} of this group, which is a circle that includes the landmark and user center positions. This is influenced by Kendon \etal \cite{kendon1990conducting} on how people form groups in interactive settings. Every sampled position $p$ has a score of
\[
Score(p) = 1.0 - Cost_{vis}(p) - Cost_{obs}(p),
\]
where we define the costs as:
\begin{align} 
\begin{split}
Cost_{vis}(p)&=(dist(p,LM)-dist(p,usr))/dist(usr,LM), \\
Cost_{obs}(p)&=max(local\_cost(p),global\_cost(p)),
\end{split}
\end{align}
where dist() is a function that returns the Euclidean distance between two 2D points, local$\_$cost(p) and global$\_$cost(p) are the cost values calculated at point p from the local and global costmap, respectively.

\begin{figure*}[h!]
\centering
        \subfigure[]{           
           \label{fig:fig16a}
           \includegraphics[trim={0 2cm 0 3cm}, clip, width=0.445\textwidth]{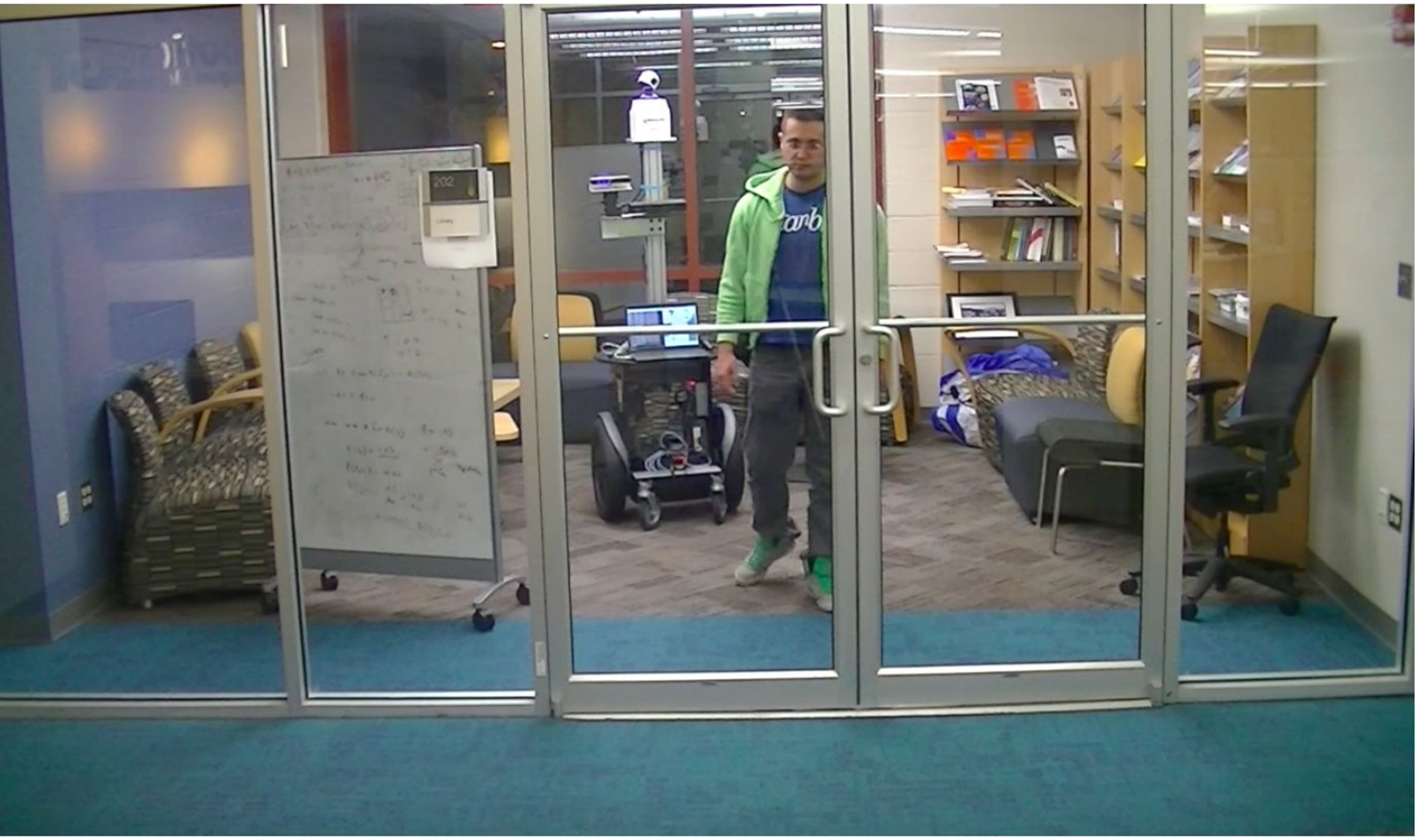}
        } 
        \subfigure[]{           
           \label{fig:fig16b}
           \includegraphics[trim={0 2cm 0 3cm}, clip, width=0.45\textwidth]{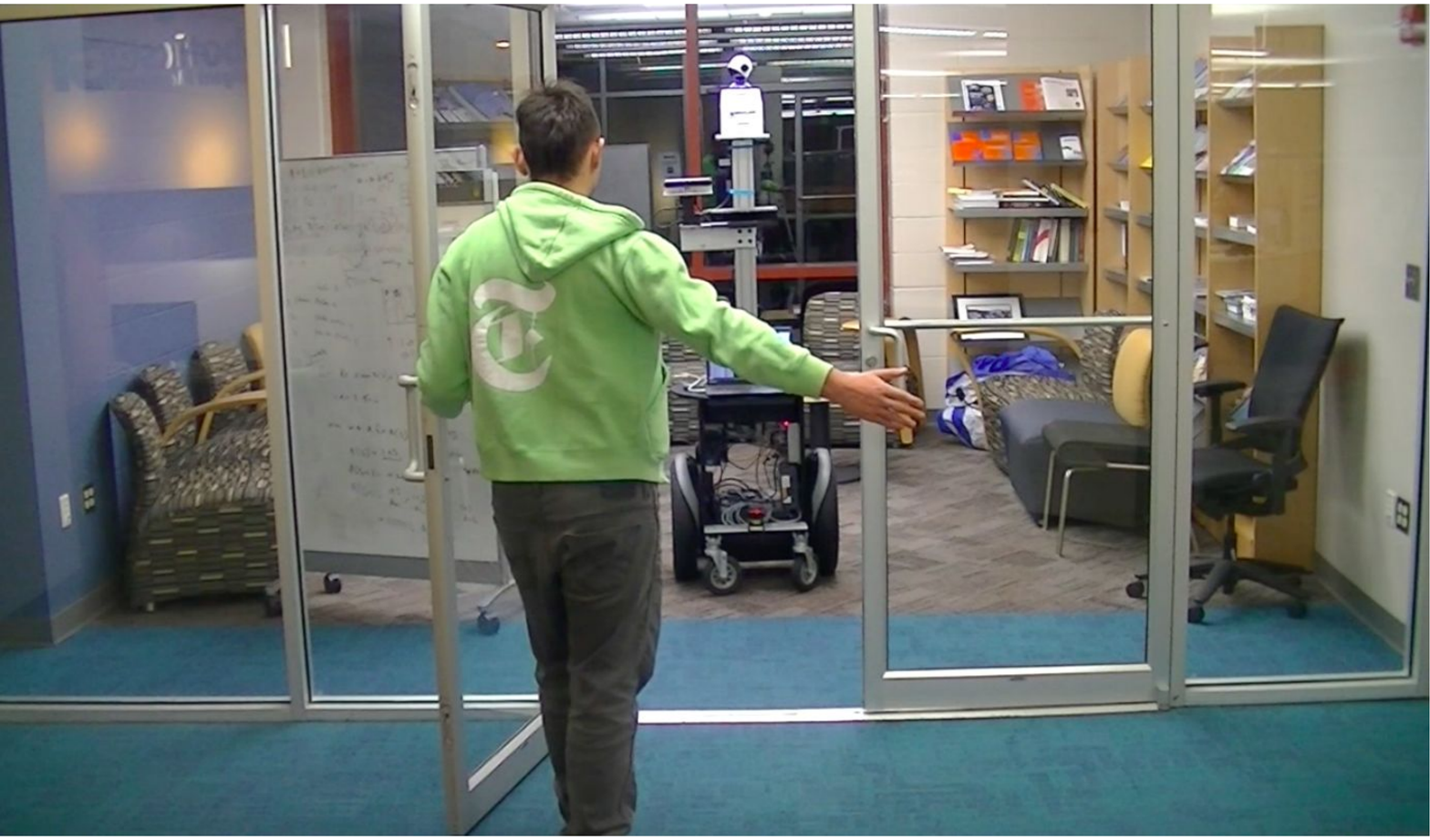}
        } \\
        \subfigure[]{
        	\label{fig:fig16c}
            \includegraphics[trim={0 2cm 0 3cm}, clip, width=0.45\textwidth]{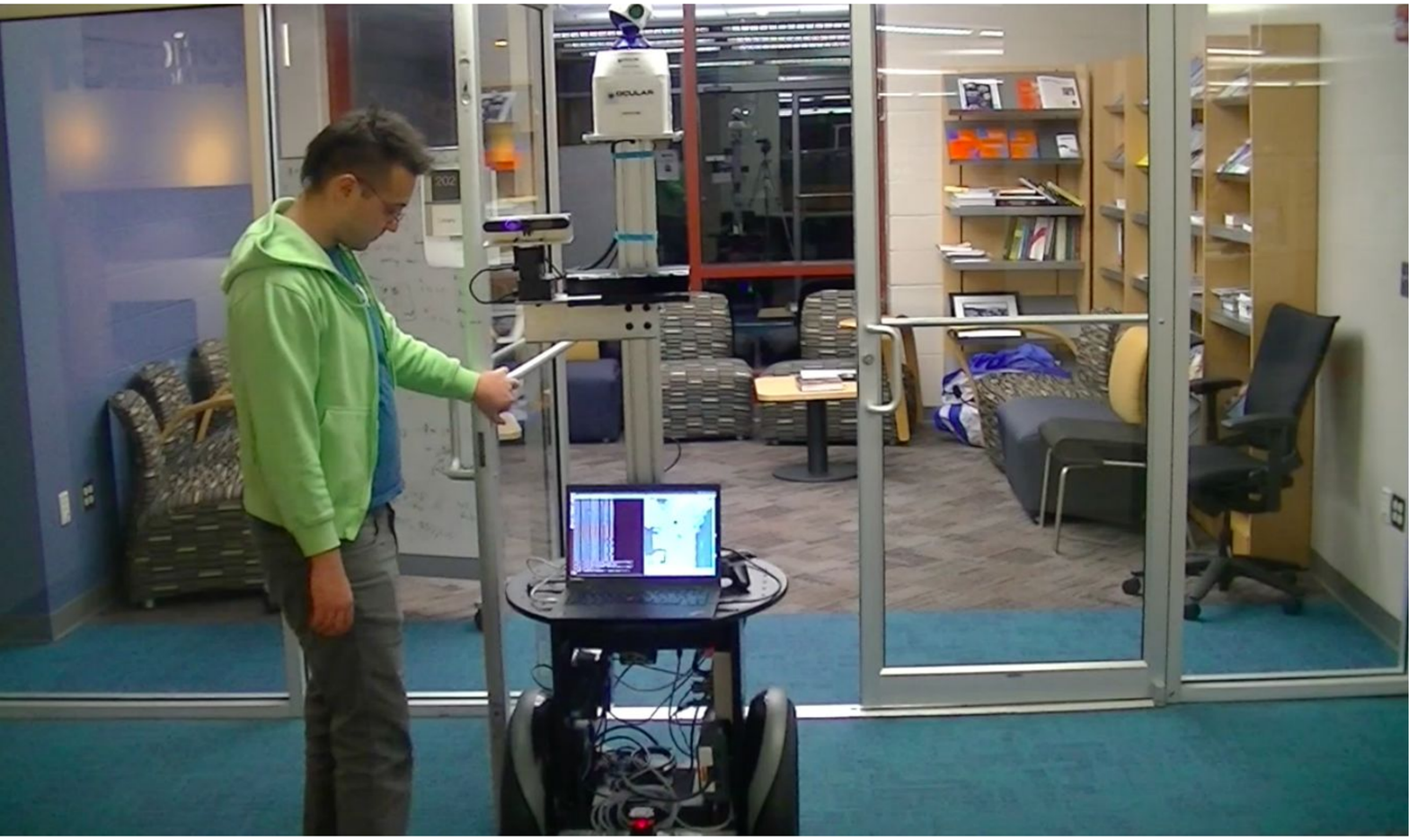}
        }        
        \subfigure[]{          
           \label{fig:fig16d}
           \includegraphics[trim={0 2cm 0 3cm}, clip, width=0.46\textwidth]{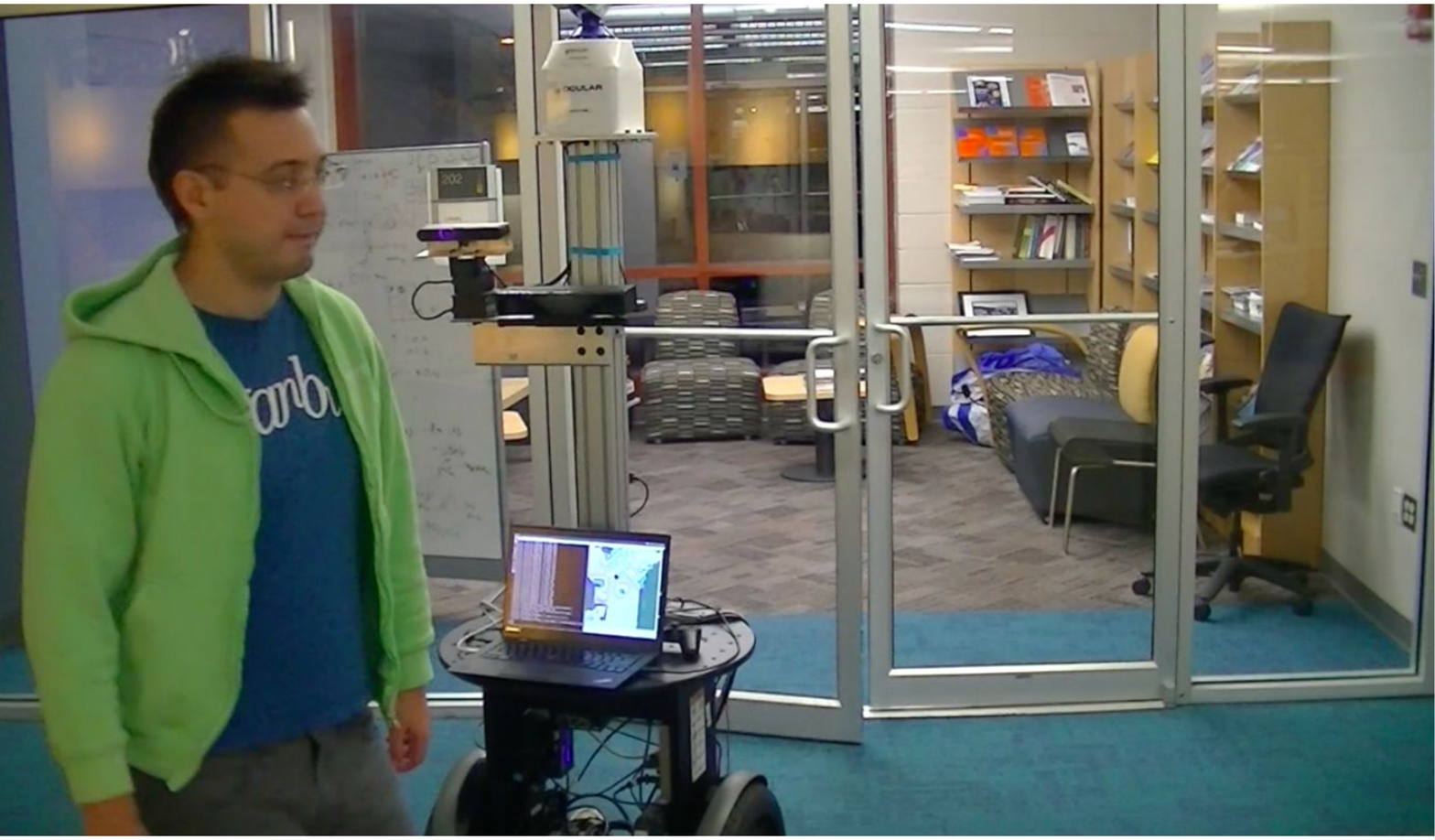}
        }
    \caption{Demonstration of context-awareness for door passing during person following. This is a swing door with spring loaded hinges, so it would close if not kept open actively. a) The robot is following the user by keeping a fixed distance to the user; b) Signal Phase: The user has stopped, is in close proximity to the door and performed a pointing gesture toward the other room; c) Approach Phase: The robot passes the door while the user is holding the door; d) Release Phase: The user has more than a threshold distance to the door, and the robot continues with the basic following.}
   \label{fig:fig16}
\end{figure*}

\begin{table*}[h!]
	\centering
  \begin{tabular}{l |  m{10cm}}    
    %\toprule    
    Signal & {dist(user, door)$<$threshold}\\         
    	      & {speed(user)$\sim$0} \\
	      & {User performs pointing gesture towards the passage}\\ \hline	                           
    Approach & {Optimal Goal: A position on the other side of the door that doesn't block the doorway}\\       \hline
    Execution & {Robot and user meet at the same side of the door}\\  \hline
    Release & {dist(user, door)$>$threshold }\\ 
    %\bottomrule
  \end{tabular}
    \caption{Conditions to trigger phases when the user is passing through a door during following.}
    \label{table:situation_aware_list_door}
    \vspace{-0.35cm}
\end{table*}

The local and global costs are fetched from the normalized local costmap which is formed by the laser scanner readings. The sample with the highest non-negative score is chosen as the goal position. The orientation of the robot is chosen as looking toward the center of all the people in the group. 

When the robot completes its move to the goal position, the user labels the landmark or objects via pointing gestures. After the task is completed, the robot waits until the user leaves the vicinity of the landmark. When that happens, the robot continues following the user. If, during any of the phases, the person tracking fails, it informs the user so following can be restarted. The phases and conditions for this behavior are summarized in Table \ref{table:situation_aware_list_landmark}. Images from a demonstration for this behavior is shown in Figure \ref{fig:fig15}.

\subsubsection{Door passing}
\label{sec:following_door_passing}

The second behavior we inspect during person following is door passing. In our experience, the reactive person following behavior can cause problems while passing doors. For example, if the user intends to close an open door or open a closed door, the robot might end up blocking the movement of the door. Moreover, a deadlock situation occurs when the user wants to go through a door with spring-loaded hinges. In that case, the user would need to hold to door to keep it open, and because the distance between the robot and the user is less than the following threshold, the robot would stay still and won't pass the door.

The robot can assume that the user might be intending to open, close or pass through a door when the user is approaching the door. In our approach, the robot continuously monitors the user's proximity to the doors using the semantic map if the door signs were detected and added to the semantic map beforehand as explained in Section \ref{sec:door_signs}. The distance check between the user and each door sign is executed each iteration by projecting the centroid of the door sign feature to the ground floor.

The phases and conditions for door passing situation are summarized in Table \ref{table:situation_aware_list_door}. The robot takes action when the user is nearby a door and performs a pointing gesture towards it to signal that the robot should pass from the door (Signal Phase). If the action is not signaled, the robot continues with basic following during the door passage. After the detection of a pointing gesture, a goal position is calculated (Approach Phase). The goal positions are sampled on the other side of the door indicated by the pointing gesture ray. A collision-free position with the least obstacle cost sample is chosen as the goal point. Note that while the robot is moving, it does not aim to keep fixed distance to the user anymore. After the robot reaches the goal, it waits for the person to pass the door (Execution Phase). After the user moved away from the door, the standard following behavior takes over. Images from the demonstration of context-aware person following for door passing is shown in Figure \ref{fig:fig16}.

There are several limitations in this work that are worth mentioning. First, we don't detect whether the door signs are to the left or right of the door and rely on the metric map to when we are sampling points for the robot. Second, when sampling goal points around the door, we make assumptions about the maximum size of the door and use a simple distance heuristic for sampling. Third, we rely on the correctness of the pointing gesture direction to figure out when sampling points from the other side of the door. Finally, we don't currently handle cases when multiple nearby doors are involved, but a data association step would be needed for those cases.

Currently, our approach to door passing only applies to doors with door signs, however, it would be possible to detect doors as planar features and label them as a door category instead of an object or planar surface.

\section{Conclusion}
\label{sec:conclusions}

In this paper we discussed the process of building semantic maps, how to interactively label entities in them, and use them to enable new navigation behaviors for specific scenarios. We utilize planar surfaces, such as walls and tables, and static objects, such as door signs as features to our semantic SLAM approach. Users can interactively annotate these features by having the robot follow him/her, entering the label through a mobile app and performing a pointing gesture toward the landmark of interest. These landmarks can later be used to generate context-aware motions.

Our pointing gesture approach can reliably estimate the target object using human joint positions and detect ambiguous gestures with probabilistic modeling. Our person following algorithm maximizes future utility by searching future actions, assuming constant velocity model for the human. We showed that our person following method can keep a near-constant distance to the human. We described a simple method to extract metric goals from a semantic map landmark and presented a human-aware path planner that considers the personal spaces of people to generate socially-aware paths. Finally, we demonstrated context-awareness for person following in two scenarios: interactive labeling and door passing. For interactive labeling, the robot utilizes the task knowledge and moves to a favorable position to facilitate interaction if an unlabeled landmark is detected near the person. For door passing, the robot utilizes the existence of a door passage by querying detected door signs in the semantic map and to execute a door passage behavior.

Semantic maps would facilitate communication of goals from an HRI perspective and enable navigation behaviors that are not feasible with metric maps. We showed proof of concept for enabling context-aware navigation behaviors using semantics and believe that there is much to explore in this research area. We think as the sensing technology improves and maps with richer semantic information become common, it would make intelligent navigation algorithms possible.

One limitation of our work is that there is only implicit interaction between the robot and human in our interaction design. The robot signaled its intention only through motion and did not explicitly communicate with people. As future work, dialogue and gaze could be utilized to complement the motions of the robot.

We think implementation and qualitative validation of robot behavior is a critical first step for path planning algorithms among humans. In this paper, we showed that our approach produced sound solutions in a number of example scenarios. As future work, we think effectiveness of context-aware navigation could be evaluated with usability studies with users who are not familiar with the robot. Moreover, scenarios could be performed under different conditions to test the generality and validate the robustness of the system.

\bibliographystyle{abbrv_mod}
\bibliography{references}

\end{document}